\documentclass[10pt,journal,compsoc]{IEEEtran}

\usepackage{amsfonts}
\usepackage{amsmath}
\usepackage{booktabs}
\usepackage[multiple]{footmisc}
\usepackage{graphicx}
\usepackage{hyperref}
\usepackage{mathtools}
\usepackage{multirow}
\usepackage{tabularx}
\usepackage[table]{xcolor}
\usepackage{subfigure}
\usepackage {makecell}

\begin{document}
\title{Differentiable Rendering: A Survey}

\author{Hiroharu Kato, Deniz Beker, Mihai Morariu, Takahiro Ando, Toru Matsuoka, \\ Wadim Kehl and Adrien Gaidon
\IEEEcompsocitemizethanks{
    \IEEEcompsocthanksitem Hiroharu Kato, Deniz Beker, Mihai Morariu, Takahiro Ando and Toru Matsuoka are with Preferred Networks, Inc. E-mail: \{hkato, dbeker, mmorariu, ando, \mbox{tmatsuoka}\}@preferred.jp
    \IEEEcompsocthanksitem Wadim Kehl is with Toyota Research Institute - Advanced Development. E-mail: wadim.kehl@tri-ad.global
    \IEEEcompsocthanksitem Adrien Gaidon is with Toyota Research Institute. E-mail: adrien.gaidon@tri.global}%
}

\markboth{Journal of \LaTeX\ Class Files,~Vol.~14, No.~8, August~2015}%
{Shell \MakeLowercase{\textit{et al.}}: Bare Advanced Demo of IEEEtran.cls for IEEE Computer Society Journals}

\newcommand{\etal}{\textit{et al}.}
\newcommand{\ie}{\textit{i}.\textit{e}.}
\newcommand{\eg}{\textit{e}.\textit{g}.}
\newcommand{\deniz}[1]{\textcolor{red}{#1}}
\newcommand{\hkato}[1]{\textcolor{magenta}{#1}}
\newcommand{\mihai}[1]{\textcolor{blue}{#1}}
\newcommand{\tmatsuoka}[1]{\textcolor{cyan}{#1}}
\newcommand{\wadim}[1]{\textcolor{purple}{#1}}
\newcommand{\ando}[1]{\textcolor{orange}{#1}}
\newcommand{\argmax}{\mathop{\rm argmax}\limits}
\newcommand{\argmin}{\mathop{\rm argmin}\limits}
\newcommand{\todo}[1]{\textcolor{magenta}{[TODO: #1]}}
\newcolumntype{L}{>{\raggedright\arraybackslash}X}

\IEEEtitleabstractindextext{\begin{abstract}
\label{sec:abstract}
Deep neural networks (DNNs) have shown remarkable performance improvements on vision-related tasks such as object detection or image segmentation. Despite their success, they generally lack the understanding of 3D objects which form the image, as it is not always possible to collect 3D information about the scene or to easily annotate it. Differentiable rendering is a novel field which allows the gradients of 3D objects to be calculated and propagated through images. It also reduces the requirement of 3D data collection and annotation, while enabling higher success rate in various applications. This paper reviews existing literature and discusses the current state of differentiable rendering, its applications and open research problems.
\end{abstract}

\begin{IEEEkeywords}
Differentiable Rendering, Inverse Graphics, Analysis-by-Synthesis
\end{IEEEkeywords}}

\maketitle

\IEEEdisplaynontitleabstractindextext
\IEEEpeerreviewmaketitle

\section{Introduction}
\label{sec:introduction}

The last years have clearly shown that neural networks are effective for 2D and 3D reasoning~\cite{krizhevsky2012imagenet,simonyan2014very,he2016deep, choy20163d,qi2016pointnet,zhou2018voxelnet}. However, most 3D estimation methods rely on supervised training regimes and costly annotations, which makes the collection of all properties of 3D observations challenging. Hence, there have been recent efforts towards leveraging easier-to-obtain 2D information and differing levels of supervision for 3D scene understanding. One of the approaches is integrating graphical rendering processes into neural network pipelines. This allows transforming and incorporating 3D estimates into 2D image level evidence.

Rendering in computer graphics is the process of generating images of 3D scenes defined by geometry, materials, scene lights and camera properties. Rendering is a complex process and its differentiation is not uniquely defined, which prevents straightforward integration into neural networks.

\begin{figure*}[t!]
    \begin{center}
        \includegraphics[width=0.9\linewidth]{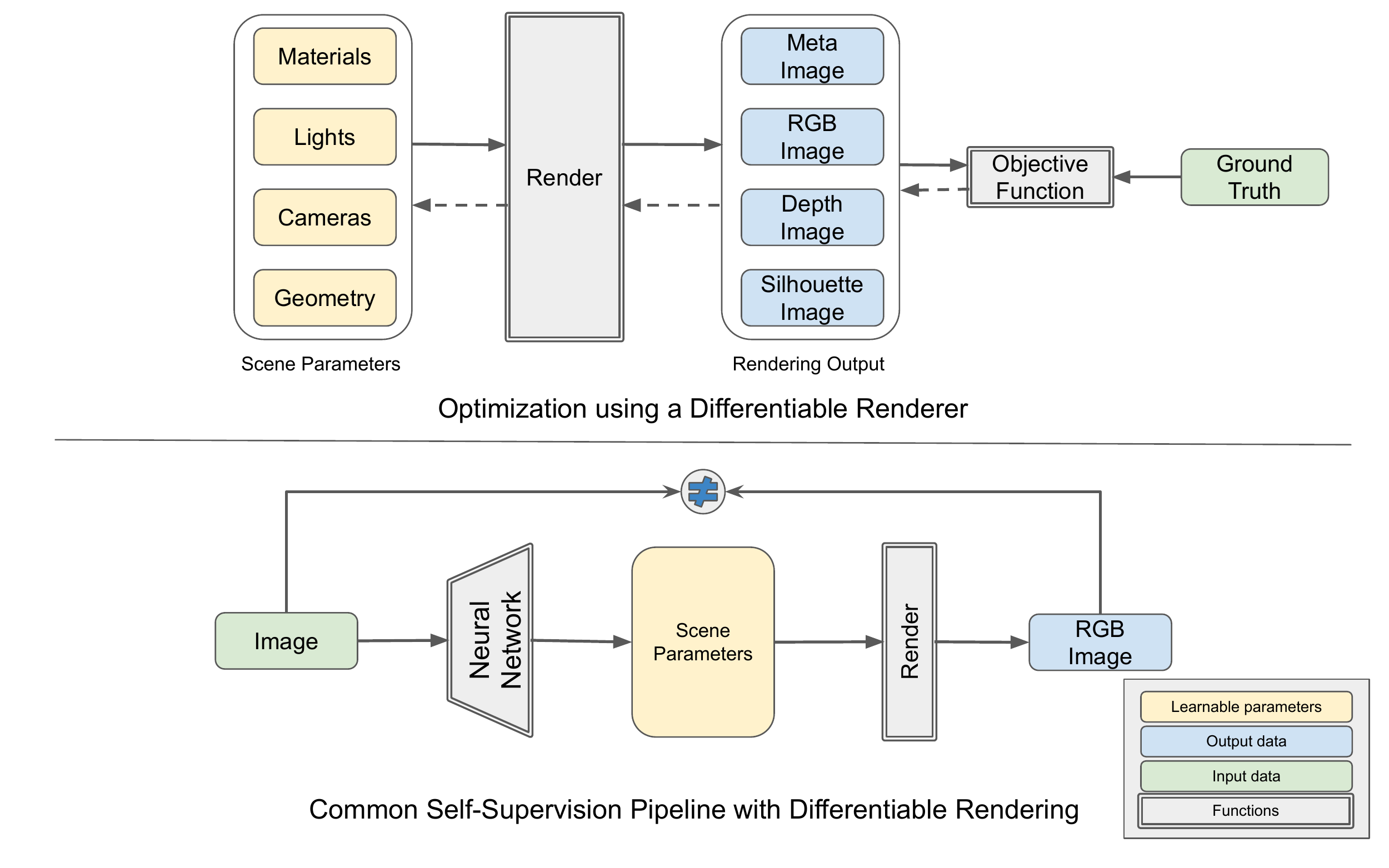}
        \label{figure:overview_of_differentiable_rendering}
        \vspace{-3mm}
        \caption{
        Schematic overview of differentiable rendering. Best viewed in color. The top part shows a basic optimization pipeline using a differentiable renderer where the gradients of an objective function with respect to the scene parameters and known ground-truth are calculated. The bottom part shows a common self-supervision pipeline based on differentiable rendering. Here, the supervision signal is provided in the form of image evidence and the neural network is updated by backpropagating the error between the image and the rendering output.}
        \vspace{-3mm}
    \end{center}
\end{figure*}

Differentiable rendering (DR) constitutes a family of techniques that tackle such an integration for end-to-end optimization by obtaining useful gradients of the rendering process. By differentiating the rendering, DR bridges the gap between 2D and 3D processing methods, allowing neural networks to optimize 3D entities while operating on 2D projections. As shown in Figure~\ref{figure:overview_of_differentiable_rendering}, optimization of the 3D scene parameters can be achieved by backpropagating the gradients with respect to the rendering output. The common 3D self-supervision pipeline is applied by integrating the rendering layer to the predicted scene parameters and applying the loss by comparing the rendered and input image in various ways. The applications of this process are broad, including image-based training of 3D object reconstruction~\cite{yan2016perspective,tulsiani2017multi,kato2018neural}, human pose estimation~\cite{pavlakos2018learning,bogo2016keep}, hand pose estimation~\cite{ge20193d,baek2019pushing} and face reconstruction~\cite{Genova_2018_CVPR,bao2019high}. 

Despite its potential, using existing or developing new DR methods is not straightforward. This can be attributed to four reasons:
\begin{itemize}
    \item Many DR-based methods have been published over the past few years. To understand which ones are suitable for addressing certain types of problems, a thorough understanding of the methods' mechanisms as well as the underlying properties is required.
    \item In order to choose or develop a novel DR method, the evaluation methodology of the existing methods should be known.
    \item In order to use DR in a novel task, it is necessary to survey the usage of DR in existing applications. However, due to the variety of applications, it is not trivial to have a clear view of the field.
    \item Several DR libraries have emerged over the past years with each focusing on different aspects of the differentiable rendering process. This makes some of the libraries useful for a particular type of applications, while for others, additional functionality may need to be implemented from scratch. Also, certain applications are constrained by computational requirements that existing implementations of DR-based methods often do not fulfill. This is especially the case for real-time applications or embedded devices, for which highly optimized neural networks are necessary.
\end{itemize}
To address these shortcomings, the current state of DR needs to be properly surveyed. However, to the best of our knowledge, there has been no comprehensive review for this purpose to date.

In this work, we provide an overview of the current state of DR algorithms in Section~\ref{sec:algorithms}, the evaluation metrics in Section~\ref{sec:evaluation}, the applications that use differentiable rendering in Section~\ref{sec:applications} and the libraries that are currently being used to facilitate the research in Section~\ref{sec:libraries}. Besides offering a survey on the existing methods, we also discuss open research problems and provide suggestions for future work.

\section{Algorithms}
\label{sec:algorithms}

\begin{table}[ht]
    \small
    \caption{Overview of the representative differentiable rendering methods. They are classified by the four main underlying data representations.}
    \centering
    \begin{tabularx}{\linewidth}{llL}
        \toprule
        Data Repr & Type & Literature \\
        \midrule
        \multirow{5}*{Mesh} & Analytical derivative & \cite{liu2018paparazzi,azinovic2019cvpr,tsai2019beyond} \\
        & Approx. gradient & \cite{loper2014opendr,kato2018neural,kato2019learning, Genova_2018_CVPR}\\
        & Approx. rendering & \cite{Rhodin:2015, liu2019soft,chen2019dibrender} \\
        & Global illumination & \cite{li2018differentiable,zhang2019differential,loubet2019reparameterizing,Zhang:2020:PSDR,NimierDavid2020Radiative} \\
        \midrule
        \multirow{2}*{Voxel} & Occupancy & \cite{yan2016perspective,drcTulsiani19, henzler2019escaping, lombardi2019neural} \\
        & SDF & \cite{jiang2019sdfdiff}  \\
        \midrule
        \multirow{3}*{Point cloud} & Point cloud & \cite{roveri2018pointpronets, yifan2019differentiable, wiles2019synsin, lin2018learning, li2020end, insafutdinov2018unsupervised} \\
        & RGBD image & \cite{godard2017unsupervised,zhou2017unsupervised} \\
        \midrule
        \multirow{2}*{Implicit} & Occupancy & \cite{liu2019learning,mildenhall2020nerf} \\
        & Level set & \cite{niemeyer2019differentiable,yariv2020universal,liu2019dist,zakharov2019autolabeling} \\
        \bottomrule
    \end{tabularx}
    \label{table:methods:papers}    
\end{table}

We start by briefly defining a mathematical formulation for our purposes. A rendering function $\mathcal{R}$ takes shape parameters $\Phi_s$, camera parameters $\Phi_c$, material parameters $\Phi_m$ and lighting parameters $\Phi_l$ as input and outputs an RGB image $I_c$ or a depth image $I_d$. We denote the inputs as $\Phi = \{ \Phi_s, \Phi_m, \Phi_c, \Phi_l \}$ and the outputs as $I = \{ I_c, I_d \}$. Note that a general DR formulation can have different kinds of additional input/output entities, but in this section we refer to the most common ones. A differentiable renderer computes the gradients of the output image with respect to the input parameters $\partial I / \partial \Phi$ in order to optimize for a specific goal (loss function). The computation of these gradients can be approximate, but should be accurate enough to propagate meaningful information required to minimize the objective function.

The inputs to a rendering function $\Phi$, especially the geometric parameters $\Phi_s$, are some of the main differentiators between the differentiable rendering algorithms. Each data representation has its own strengths which makes it suitable for solving specific problems. In this work we group the surveyed DR algorithms into four categories, based on the underlying data representation: mesh, voxels, point clouds and neural implicit representations. We also discuss \textit{neural rendering}, as there is a growing body of research into learning to render using neural network models, rather than designing the rendering and its differentiation manually. Table~\ref{table:methods:papers} lists the literature that we are covering in this section.

\begin{figure}[t]
    \begin{center}
        \includegraphics[width=1.0\linewidth]{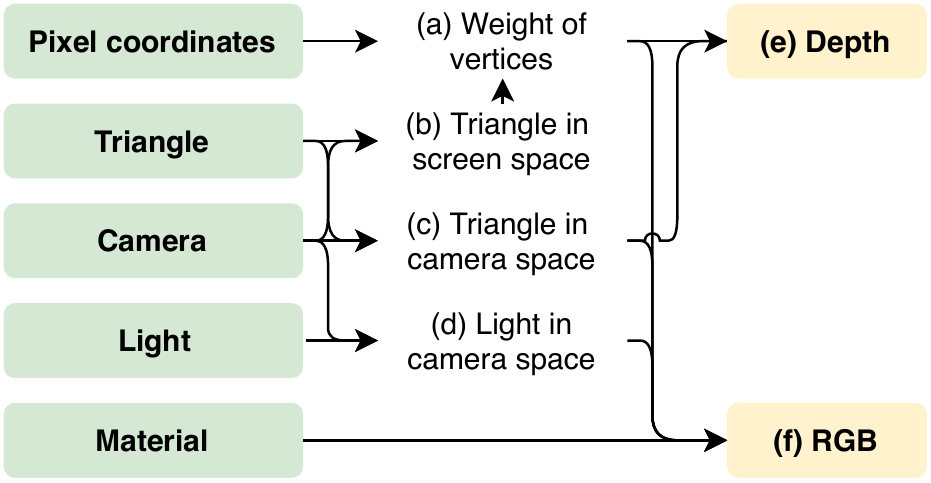}
        \caption{Several operations that are performed inside a rendering function, given a pixel, its corresponding triangle and material defined on vertices of the triangle, camera parameters, and light configurations. The green boxes represent inputs and the yellow boxes represent outputs. Best viewed in color.}
        \label{figure:rendering_detail}
    \end{center}
\end{figure}

\begin{figure}[t]
    \begin{center}
        \includegraphics[width=0.8\linewidth]{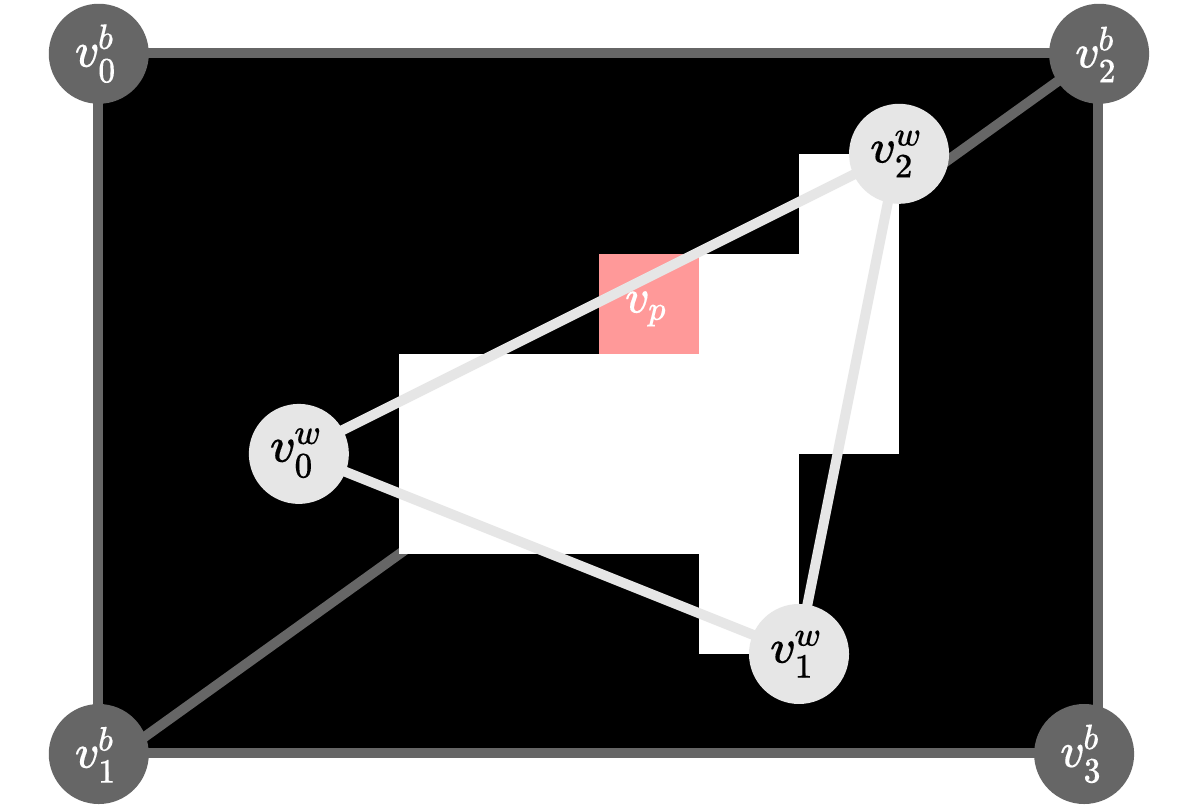}
        \caption{An image of $10 \times 7$ pixels that shows a scene composed of three triangles. The vertex colors of one are white and its vertex positions are denoted by $v_i^w$. The vertex colors of the other two are black and their vertex positions are denoted by $v_i^b$. }
        \label{figure:rendering_pixel}
    \end{center}
\end{figure}

\subsection{Mesh}
\label{sec:algorithms:mesh}

\subsubsection{Analytical Derivative of Rendering}

A mesh represents a 3D shape as a set of vertices and the surfaces that connect them. It is widely used, especially in computer graphics, because it can represent complex 3D shapes in a compact way. 

Given the rendering inputs and a pixel, the process of determining its color can be divided into (1) assigning a triangle to the pixel and (2) computing the color of the pixel based on the colors of the assigned triangle's vertices. The former is done by projecting all mesh triangles from world space to screen space and determining the ones that enclose the pixel. The triangle that is closest to the camera is then selected. Since the selection yields a discrete triangle identifier, this operation is not differentiable with respect to all parameters.

All further operations are differentiable. Figure~\ref{figure:rendering_detail} illustrates the simplified computation flow. First, the triangle, the position and direction of the lights are projected from world space to camera space (Figure~\ref{figure:rendering_detail} (c), (d)), followed by a transformation into screen space (Figure~\ref{figure:rendering_detail} (b)). These operations are differentiable because they are accomplished via simple matrix products. The pixel coordinates can then be expressed as a weighted sum of three vertices. The computation of the weights (Figure~\ref{figure:rendering_detail} (a)) is done by solving (differentiable) linear programs whereas the pixel depth is computed by interpolating the depth of the three vertices in camera space (Figure~\ref{figure:rendering_detail} (d)). Similarly, the material and normal vectors at a pixel are typically expressed as a weighted sum of the material and normal vectors defined at the triangle's vertices. For local illumination models, given the material and lighting parameters, as well as the normal vector at a pixel, the pixel color can be computed using a reflection model (Figure~\ref{figure:rendering_detail} (e)). Popular reflection models such as Phong~\cite{phong77}, Lambertian~\cite{lambert} and Spherical Harmonics~\cite{spherical_harmonics} are all differentiable. Therefore, the derivatives of the pixel depth and color with respect to the input parameters can be computed analytically~\cite{liu2018paparazzi,azinovic2019cvpr,tsai2019beyond}.

In standard rendering, exactly one triangle per pixel is typically selected to compute the final color value, which can lead to optimization problems. To exemplify, let us consider a scene composed of one white triangle and two black triangles as illustrated in Figure~\ref{figure:rendering_pixel}. The vertex color of $v_i^w$ is 1 and the vertex color of $v_i^b$ is 0. Then, using barycentric coordinates $w_i$ that satisfy $v_p = \sum w_i v_i^w$ and $\sum w_i = 1$, the color of a pixel at $v_p$ is a constant $c_{v_p} = \sum w_i c_{v_i^w} = 1$. Therefore, the gradient of $c_{v_p}$ with respect to $v_i^w$ and $v_i^b$ is zero.

Similarly, the derivative of a pixel color with respect to a vertex position is always zero for all pixels and vertices. Therefore, analytical derivatives do not help with optimizing the geometry in this case. However, in practice, the position of vertices affects pixel colors. For example, when $v_2^w$ moves to the right, $c_{v_p}$ would change to 0. We can therefore solve the problem by allowing the color of unrelated pixels to affect neighboring triangles. Several rendering methods provide approximated gradients that reflect these insights~\cite{loper2014opendr,kato2018neural,kato2019learning,Genova_2018_CVPR}, while others overcome this problem by approximating the rasterizer pass~\cite{liu2019soft,chen2019dibrender,Rhodin:2015}.

\subsubsection{Approximated Gradients}
\label{sec:algorithms:mesh:approximated-gradient}

Loper and Black~\cite{loper2014opendr} employ approximated spatial gradients in the first general purpose differentiable renderer, named {\it OpenDR}. $v_p$ can be represented by $v_p = \sum w_i v_i^w$ with $\sum w_i = 1$ and the pixel derivatives with respect to $v_p$ can be computed using differential filters (e.g. Sobel filter). In other words, $\{ \frac{\partial c_{v_p}}{\partial x}, \frac{\partial c_{v_p}}{\partial y} \} = \frac{\partial c_{v_p}}{\partial v_p}$. Because the pixels that are located to the left and right of $v_p$ are taken into consideration when computing the gradient, it can have a non-zero value in this formulation.

Kato \etal~\cite{kato2018neural} raise two issues with OpenDR and propose a renderer named {\it neural 3D mesh renderer (NMR)}. The first issue is the localness of gradient computation. Because of the localness of differential filters in OpenDR, only gradients on boundary pixels can flow towards vertices, whereas gradients at other pixels cannot be used. Optimization based on this property may lead to poor local minima. The second issue is the derivative does not leverage the loss gradient of the target application, e.g. image reconstruction. For example, in the case of Figure~\ref{figure:rendering_pixel}, if the objective is to decrease the intensity of $v_p$, we should displace $v_2^w$. However, if the objective is to increase it, we should not. Therefore, gradients should be objective-aware for better optimization. Since the objective of OpenDR is not to provide accurate gradients but to provide useful gradients for optimization, a loss-aware gradient flow is required for this purpose. To overcome these issues, the authors propose non-local approximated gradients that also use gradients of pixels backpropagated from a loss function. The authors later replaced non-local gradients with local gradients similar to OpenDR, in order to reduce the computation complexity~\cite{kato2019learning}.

Genova \etal~\cite{Genova_2018_CVPR} calculate the rasterization derivatives using the barycentric coordinates of each triangle with respect to each pixel. They introduce negative values for the barycentric coordinates of the pixels that lie outside the triangle border, in order to overcome the occlusion discontinuity. By omitting triangle identifiers and by employing negative barycentric coordinates, shapes can be treated as being locally planar, to approximate the occlusion boundaries. However, such approximation could pose problems when optimizing for translation or occlusion.

\subsubsection{Approximated Rendering}
Instead of approximating the backward pass, other methods approximate the rasterization (or the forward pass) of the rendering, in order to be able to compute useful gradients.

Rhodin \etal~\cite{Rhodin:2015} reinterpret the scene parameters to ensure differentiability. To prevent the discontinuity at hard object boundaries, each object is defined by a density parameter which has the maximum opaqueness at the object's center and becomes transparent towards the boundaries. As a result, the rendering result is blurry and smooth towards the edges, while removing sharp corners from the scene parameters ensures differentiability. 

Liu \etal~\cite{liu2019soft} take a similar approach and propose a renderer named {\it Soft Rasterizer}. In addition to spatial blurring, it replaces the z-buffer-based triangle selection of a vanilla rasterization process with a probabilistic approach in which each triangle that is projected onto a pixel $p_i$ contributes to its color with a certain probability. In practice, an aggregation function fuses all the color probabilities for each pixel. As a result, each pixel color is computed as a weighted sum of the values corresponding to the relevant triangles and that operation is differentiable. The weights are based on the distances between a pixel and a triangle in the 2D screen space, as well as the distance between the camera and the triangle, along the viewing direction. Therefore, gradients accumulate information across the whole image in a probabilistic way, while OpenDR only backpropagates to a vertex from neighbouring pixels and NMR only backpropagates gradients for the pixels that lie within $ [min(obj_x), max(obj_x)]$ and $[min(obj_y),  max(obj_y)]$. Note that all methods in the previous section provide no control over the forward pass as they aim to approximate the backward gradients only. 

Chen \etal~\cite{chen2019dibrender} propose {\it DIB-R}, which focuses independently on two different regions of the image: foreground pixels, where a pixel is covered by at least one face, and background pixels, which do not have any face coverage. To avoid the blurry output of Soft Rasterizer, DIB-R proposes to use analytical derivatives for foreground pixels, computed using the barycentric interpolation of a face's vertex attributes. It also prevents vanishing gradients for background pixels by employing a distance-based aggregation of global face information in a similar way to Soft Rasterizer.

\subsubsection{Global Illumination}
The pipeline in Figure~\ref{figure:rendering_detail} does not hold for global illumination models because lighting for a pixel is not affected by reflected light from other surface points. Although this simplification reduces the rendering time, the generation of photorealistic images that contain complex interactions of light, geometry, and materials becomes impossible. The color of a pixel is computed using the Monte Carlo estimation of the rendering equation~\cite{kajiya1986rendering} in global illumination models. The main challenge in differentiable photorealistic rendering is estimating the derivative corresponding to the integral of the Monte Carlo-estimated rendering equation when the integral contains discontinuities due to object silhouettes.

Li \etal~\cite{li2018differentiable} is the first work to compute derivatives of scalar functions over a physically-based rendered image with respect to arbitrary input parameters like camera, light materials and geometry. It uses a stochastic approach based on Monte Carlo ray tracing which estimates both the integral and the gradient of the pixel filter's integral. As edges and occlusions are discontinuous by nature, the integral calculation is split into smooth and discontinuous regions. For the smooth parts of the integrand, a traditional area sampling with automatic differentiation is employed. For the discontinuous parts, a novel edge sampling method is introduced to capture the changes at boundaries. Their method makes certain assumptions: meshes do not have interpenetration, there are no point light sources, no perfectly specular surfaces and the scene is static. Zhang \etal~\cite{zhang2019differential} propose a very similar method. Different from Li \etal~\cite{li2018differentiable}, their approach supports the differentiation of volumes in addition to triangle meshes. Two major drawbacks of these methods are the rendering speed and the large variance of the estimated gradients. This is due to the challenging task of finding all object edges and sampling them, for which many samples are required.

Instead of relying on edge sampling, Loubet \etal~\cite{loubet2019reparameterizing} propose to reparametrize all relevant integrals, including pixel integrals over spherical domains. Discontinuity occurs at those points that depend on a scene parameter, therefore they reparametrize the variables from the integrals to remove this dependency. Such reparametrizations allow integration over a space where discontinuity does not take place when a scene parameter changes and is equivalent to importance sampling the integral that follows the discontinuity. Even though this approach is computationally efficient, it does not support perfectly specular materials, degenerate light sources containing Dirac delta functions and multiple discontinuities within the support of the integrand. In addition, since the gradients are approximated, they may not always be accurate.

Zhang \etal~\cite{Zhang:2020:PSDR} propose a method to estimate the derivatives of the path integral formulation~\cite{veach1997robust} while all the methods reviewed above address the discontinuity problem in the rendering equation~\cite{kajiya1986rendering}. The authors show that the differentiation of path integrals can be separated into {\it interior} and {\it boundary} terms and propose Monte Carlo methods for estimating both components. The proposed method is unbiased and computationally efficient as it does not need to find object silhouette edges explicitly. However, because the computation of the gradients for a single rendered image takes anywhere between a few second to tens of seconds, it is impractical for training a neural network.

The gradient with respect to material and light parameters can be computed by automatic differentiation. However, given its large memory footprint, applications are limited to simple scenes. To address this issue, Nimier-David \etal \cite{NimierDavid2020Radiative} propose an efficient approach to gradient computation. In their method, a computational graph is not stored during rendering. Instead, during backpropagation, rays with gradients are cast from the camera and gradients are propagated to surfaces at occlusion. However, optimizing shapes using this method is challenging because a change in object visibility is not differentiable and not considered during backpropagation.
\subsection{Voxel}

In this section we survey differentiable rendering algorithms that use voxels to represent data. A voxel is a unit cube representation of a 3D space. It can be parametrized as an N-dimensional vector that contains information about the volume that is occupied in 3D space, as well as additional information. It is common practice to encode occupancy information in a voxel using a binary value or transparency using a non-binary one. For applications where occupancy is predicted, non-binary occupancy probability $P_{o} \in [p_{min}, p_{max}]$ is usually stored. Even though occupancy probabilities are different from transparency values, they can be interpreted the same way in order to maintain differentiability during the ray marching operation. In this case, the probability $P_{o}$ denotes a ray's absorption (transparency) at a certain point. Material information is often also stored. Instead of storing occupancy, shapes can be represented as the shortest distance from the center of each voxel to the object's surface. In this representation, each voxel cell is assigned a distance function (DF). Distance functions can be augmented with a signed value denoting whether the voxel is contained inside or outside the object, to form a Signed Distance Function (SDF). Alternatively, truncation can be applied to an SDF to form a Truncated Signed Distance Function (TSDF) for those applications where only the distance information to the object's surface is important.

All the voxels that are located along a ray that projects to a pixel are taken into account when rendering that pixel. Several approaches exist for deciding the resulting pixel color~\cite{yan2016perspective,drcTulsiani19,henzler2019escaping,lombardi2019neural,jiang2019sdfdiff}. Since the position of a voxel is fixed in the 3D space, the gradient issue described in Section~\ref{sec:algorithms:mesh}, which is caused by the displacement of shape primitives, does not occur when rendering voxels.

Collecting the voxels that are located along a ray is the first step of the rendering process. Tulsiani \etal~\cite{drcTulsiani19} and Jiang \etal~\cite{jiang2019sdfdiff} perform this operation in world space, while Yan \etal~\cite{yan2016perspective} and Henzler \etal~\cite{henzler2019escaping} take a different approach. They project the voxels from the world space to the screen space (using camera parameters) and perform a bilinear sampling similar to Spatial Transformer~\cite{spatial_transformer_networks}, which is more computationally efficient. The approach of Lombardi \etal~\cite{lombardi2019neural} is rather different. They introduce the notion of warp field (which is the mapping between the template volume and output volume) to improve the apparent resolution and reduce grid-like artifacts along with jagged motion. The inverse wrap value is evaluated and the template volume is sampled at the warped point. Since all the described operations are based on a subset of voxels, the output is differentiable with respect to them.

Aggregating voxels along a ray is the second step of the rendering process. Yan \etal~\cite{yan2016perspective} associate an occupancy probability to each pixel. This value is computed by casting a ray from the pixel, sampling all the associated voxels and choosing the one with the highest occupancy probability. Instead of taking the maximum value, Tulsiani \etal~\cite{drcTulsiani19} compute the probability that a ray stops at a certain distance. The advantage of this method is the ability to render depth maps and color images in addition to the foreground mask of an object. Henzler \etal~\cite{henzler2019escaping} further introduce the bidirectional emission-absorption (EA) light radiation model for the emitting material, in addition to the visual hull (VH) and absorption-only (AO) models which are similar to Tulsiani \etal. Lombardi \etal~\cite{lombardi2019neural} employ differentiable ray marching by casting a ray from each pixel and then iteratively accumulating color and opacity probabilities along it. When the cumulative sum of the opacity values along the ray reaches a maximum value, the remaining voxels are discarded in order to prevent further impact to its color. 

Different from these aforementioned methods, Jiang \etal~\cite{jiang2019sdfdiff} handle voxels that represent signed distance functions. They cast a ray from the pixel and locate the surface voxel that is closest to the camera, along the ray. The final pixel color is computed as a weighted sum of the surface voxel and its neighbouring voxels. They also compute a normal vector at the surface point, based on the voxel values around it, in order to compute a shading value. Although locating the surface point is not a differentiable operation, the color of the pixel is differentiable with respect to the voxels around the surface point.

Several works also take materials into account during the rendering process. Tulsiani \etal~\cite{drcTulsiani19}, Henzler \etal~\cite{henzler2019escaping}, and Lombardi \etal~\cite{lombardi2019neural} simply use voxel colors to represent the material. Similar to mesh rendering, the rendered image is differentiable with respect to the material parameters, since most of the shading models are differentiable.

\subsection{Point Cloud}
\label{sec:algorithms:point-cloud}

Point clouds are collections of points that represent shapes in the 3D space. They are ubiquitous in the 3D vision community due to their ability to represent a large variety of topologies at a relatively low storage cost. Moreover, most of the 3D sensors that are nowadays available on the market also rely on point clouds for encoding data. Recent years have shown that point clouds can successfully be integrated into deep neural networks to solve a variety of practical 3D problems~\cite{zhou2018voxelnet,roynard2018classification,qi2016pointnet}. Given the advent of differentiable rendering and its potential for scene understanding with reduced 3D supervision, point clouds have thus become a natural choice for data representation.

Rendering a point cloud is done in three steps. First, the screen space coordinates $p_{uv}^i$ of a 3D point $P_{xyz}^i$ is calculated. This operation is achieved by a matrix multiplication as with rendering of mesh. Therefore, this step is differentiable. Second, the influence $I_i$ of each 3D point on the color of target pixel's color is computed. Third, the points are aggregated based on their influences and z-values, in order to decide the final color value for a pixel. Several methods for addressing this step have been proposed as well.

The straightforward way to calculate the $I_i$ is to assume that $p_{uv}^i$ has a size of one pixel~\cite{lin2018learning}. However, this approach may lead to very sparse images. One way to solve this problem is creating influences larger than one pixel per $p_{uv}^i$  in the image. Several papers employ a truncated Gaussian blur~\cite{roveri2018pointpronets,yifan2019differentiable} or influence values based on the distance between a pixel and $p_{uv}^i$~\cite{wiles2019synsin,li2020end}. The generated images are differentiable with respect to the points when these operations are implemented in an autodiff framework. However, in these methods, there is a trade-off between the rendering quality and the optimization's convergence. A large influence size prevents the derivative of a pixel with respect to a distant $P_{xyz}^i$  from becoming zero, but reduces the rendering quality. To address this issue, Yifan \etal~\cite{yifan2019differentiable} propose the invisible approximated gradient, similar to Kato \etal~\cite{kato2018neural}.

Several methods have been proposed for computing the color of a pixel. A straightforward way is computing a weighted sum of the points' colors, based on the points' influences~\cite{yifan2019differentiable}. However, this approach does take occlusion into account. Lin \etal~\cite{lin2018learning} address this issue by selecting the closest point to the camera for a given pixel, which prevents the optimization of the occluded points. Instead, Li \etal~\cite{li2020end} propose to select the $K$ nearest 3D points $P_{xyz}$ to the camera and weigh them based on the spatial influence $I^i$ at each pixel. Wiles \etal~\cite{wiles2019synsin} propose to weigh all the 3D points $P_{xyz}$ according to the distances from the camera to $P_{xyz}$, in addition to weighing by spatial influence $I^i$, similar to Li \etal~\cite{liu2019soft}.

Insafutdinov \etal~\cite{insafutdinov2018unsupervised} take an unique approach. Instead of computing influence values and aggregating points in screen space, they generate 3D voxels from the point cloud and render them using a method by Tulsiani \etal~\cite{tulsiani2017multi}.

A set of depth image and camera parameters is an alternative way to represent point cloud data. One of the major benefits of using a depth image is that its spatial information can be used directly to recover information about the triangulation between points. On the other hand, when projecting a depth image to the 3D space, the point density decreases inversely proportional to the distance. Generation of another view from an depth image can be regarded as a special case of rendering of point clouds. In self-supervised learning of monocular depth estimation~\cite{godard2017unsupervised,zhou2017unsupervised}, a color image is used to estimate the corresponding depth, which is converted to a point cloud and projected to a virtual camera view.

\subsection{Implicit Representations}
\label{sec:algorithms:implicit-representations}

\begin{table*}[ht]
    \small
    \centering
    \caption{Comparison of differentiable rendering methods for voxels and neural implicit functions.}
    \begin{tabularx}{1.0\linewidth}{llll}
        \toprule
        Operation & Approach & Voxel & Neural implicit functions \\
        \midrule
        \multirow{2}*{Data collection along a ray} & Sampling along a ray & \cite{drcTulsiani19,lombardi2019neural,jiang2019sdfdiff} & \cite{mildenhall2020nerf,niemeyer2019differentiable,yariv2020universal,liu2019dist,zakharov2019autolabeling} \\
        & (Re)sampling in 3D space & \cite{yan2016perspective,henzler2019escaping} & \cite{liu2019learning} \\
        \midrule
        \multirow{3}*{Data aggregation along a ray} & Approx. occupancy: taking the maximum value & \cite{yan2016perspective} & \cite{liu2019learning} \\
         & Transparency: taking a weighted sum & \cite{drcTulsiani19,henzler2019escaping,lombardi2019neural} & \cite{mildenhall2020nerf} \\
         & Level set: taking the value at a hit point & \cite{jiang2019sdfdiff} & \cite{niemeyer2019differentiable,yariv2020universal,liu2019dist,zakharov2019autolabeling} \\
        \bottomrule
    \end{tabularx}  
    \label{table:methods:voxels_and_implicit_functions}  
\end{table*}

Recently, there has been a growing interest in representing geometric information in a parametrized manner in neural networks~\cite{chen2019learning,mescheder2019occupancy,park2019deepsdf}. This is commonly known as {\it neural implicit representation}. In this model, the geometric information at a point $P_{xyz}^i \in \mathbb{R}^3$ is described by the output of a neural network $F(P_{xyz}^i)$. Unlike voxel-based methods, in implicit representations the memory usage is constant with respect to the spatial resolution. Therefore, a surface can be reconstructed at infinite resolution without excessive memory footprint.

Similar to voxel-based methods, there are three ways to represent the geometric information. First, the probability that a point $P_{xyz}^i$ is occupied by an object can be modeled by a neural network $F$. When the ground-truth occupancy at $P_{xyz}^i$ is given, learning $F$ becomes a binary classification problem and has been extensively studied in the literature. Second, $F$ can model the transparency at a point $p$. This approach can be used for representing semi-transparent objects, as well as approximating the probability of occupancy for scenes with no semi-transparent objects. Third, the surface of an object can be defined as the set of points $P_{xyz}^i$ which satisfy $F(P_{xyz}^i) = 0$ and which is called {\it level-set method}. Typically, $F(P_{xyz}^i)$ defines the distance to the boundary of the target surface, while its sign indicates whether $P_{xyz}^i$ is inside or outside the surface.

As with other representations, obtaining ground-truth 3D shapes is often expensive or impossible, for real-world scenarios. To address this issue, several works propose to use 2D supervision (in the form of depth maps or multi-view images) and leverage the power of differentiable rendering~\cite{liu2019learning,mildenhall2020nerf,niemeyer2019differentiable,yariv2020universal,liu2019dist,zakharov2019autolabeling}. Similar to voxel-based methods, for which various differentiable rendering algorithms have been developed for approximated occupancy probability~\cite{yan2016perspective}, occupancy probability and transparency~\cite{drcTulsiani19} and implicit surfaces by distance functions~\cite{jiang2019sdfdiff}, implicit representations also require implementations to handle the different input types. Table~\ref{table:methods:voxels_and_implicit_functions} summarizes the similarities between methods based on voxels and the ones based on neural implicit functions.

Liu \etal~\cite{liu2019learning} propose the first usage of neural implicit representations in differentiable rendering. To find the occupancy probability of a pixel $p_{uv}$, a ray $R$ is cast from $p_{uv}$. Then, the 3D point $p_{xyz}$ with the maximum occupancy probability value along the ray $R$ is selected. Finally, the occupancy probability of $p_{uv}$ is assigned the value of $p_{xyz}$. The gradient value of $p_{uv}$ is assigned to $p_{xyz}$ during backpropagation. 
By contrast, Mildenhall \etal~\cite{mildenhall2020nerf} propose differentiable rendering for neural transparency. In their paper, a pixel value is computed using volume rendering and by weighing the values of all the sampled points along a ray. Therefore, gradients flow into more than one point, which is expected to stabilize optimization. Although Liu \etal~only support the rendering of silhouettes, Mildenhall \etal~support the rendering of RGB images using Texture Fields~\cite{oechsle2019texture}. Another important aspect regarding the work by Mildenhall \etal~is that they take the direction of a ray into account when computing color values and, therefore, support view-dependent phenomena such as specular reflection.

Several works~\cite{niemeyer2019differentiable,yariv2020universal,liu2019dist,zakharov2019autolabeling} that employ neural implicit functions represent surfaces by the level set method in which a ray is cast from each pixel and the intersection between the ray and the surface is used to compute the camera-to-surface distance. The color at the intersection point is sampled from a neural network. The derivative of the distance with respect to the intersection point cannot be computed by an autodiff framework, but it can be computed analytically. These methods cannot render differentiable silhouette images because the distance to the surface is infinite for background pixels. Therefore, to optimize while taking silhouettes into account, it is common to employ different loss functions for foreground and background pixels, respectively. For example, Niemeyer \etal~\cite{niemeyer2019differentiable} minimize the occupancy of the intersection point when encountering the projection to the silhouette is a false positive and maximize the occupancy of a random point along a ray when the projection is a false negative.

Sampling points on a ray is easy for voxel-based methods, as there is a finite set of voxels along it, due to the discrete nature of the 3D grid. For neural implicit representations, there is an infinite set of points along the ray and the sampling process becomes challenging. Two of the most common ways to address this issue are sampling random points or regular points with random perturbations~\cite{mildenhall2020nerf,niemeyer2019differentiable,yariv2020universal,liu2019dist,zakharov2019autolabeling}. Another approach is sampling random points in the 3D space and checking their intersection with the rays, which becomes even more efficient when multiple views are rendered~\cite{liu2019learning}. To optimize even further, many methods employ coarse-to-fine sampling on rays~\cite{mildenhall2020nerf,niemeyer2019differentiable,yariv2020universal} and importance sampling near boundaries~\cite{liu2019learning}.

\subsection{Neural Rendering}
\label{sec:algorithms:neural-representations}

Instead of handcrafting the rendering differentiation, Eslami \etal~\cite{eslami2018neural} propose to learn the rendering process from data. This approach is commonly known as {\it neural rendering}. Typically, a scene generation network that outputs a neural scene representation and a rendering network are jointly trained by minimizing the image reconstruction error. Thanks to the recent advances in neural networks, neural rendering is nowadays becoming capable of generating high-quality images and is used for many applications such as novel view synthesis, semantic photo manipulation, facial and body reenactment, relighting, free-viewpoint video and to the creation of photo-realistic avatars.

While handcrafted renderers do not always model the physical world accurately, by learning from real-world data, neural rendering can produce novel images that are indistinguishable from the real-world ones. On the other hand, generalizing to scenes that differ from the training data, scaling to scenes composed of multiple objects and the ability to be modified by humans are limited. One promising direction for improving neural rendering is adding inductive biases for 3D scenes and rendering into neural networks~\cite{sitzmann2019scene,lombardi2019neural,nguyen2019hologan}. Therefore, combining differentiable renderers which are based on inductive biases with neural rendering would be one interesting research area to explore.

Please refer to a recent survey~\cite{tewari2020state} for more details about neural rendering. 
\subsection{Summary}
\label{sec:algorithms:summary}

\begin{figure}[t]
    \begin{center}
        \includegraphics[width=\linewidth]{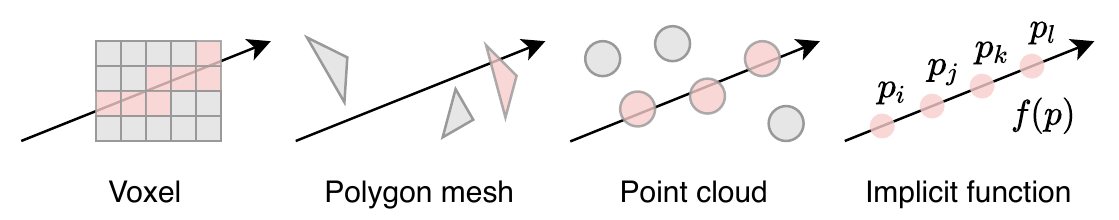}
        \caption{Differentiable rendering algorithms differ in how the geometric information along a ray is collected and aggregated. For voxels, collecting geometric information is done by checking intersections of a ray and each voxel. For meshes, unlike non-differentiable rendering, multiple polygons have to be associated with a single ray. For point cloud, there are several ways to measure the influence of a point to a ray by pseudo-sizing. For neural implicit functions, various sampling techniques have been proposed for efficiency. Aggregation methods mainly depend on whether geometric information is treated deterministically or probabilistically, and how occlusion is handled.}
        \label{figure:algorithms_representations}
    \end{center}
\end{figure}

We have presented differentiable rendering techniques grouped based on four data representation types: mesh, voxel, point cloud and implicit functions. Figure~\ref{figure:algorithms_representations} illustrates these representations. The main points of this section can be summarized as follows:

\textbf{Mesh-based approaches can be grouped into three categories: approximated gradients, approximated rendering and global illumination.} Gradient approximation allows for efficient and high-quality rasterization while aiming to handcraft meaningful gradients. Rendering approximation may produce blurry output, but ensures non-zero gradients for all pixels. Global illumination-based techniques reduce the domain gap between the rendered image and real-world data, but are currently impractical for usage with deep neural networks due to their high computation cost. \\

\textbf{ Voxel-based approaches are easy to use, but require an excessive amount of memory and parameter usage.} The collection and aggregation of voxels along the ray, to produce a final pixel value, is a major differentiating factor of voxel-based approaches. Despite the easiness and strength of such methods, their applicability is limited to small scenes and low resolutions. SDF-based voxel approaches allow representing surfaces more smoothly than occupancy grid-based approaches.

\textbf{ Point cloud-based approaches provide low computational cost while facing the sizing ambiguity.} Point clouds are a natural choice for many differentiable rendering methods due to their simplicity and widespread usage with 3D sensors, but fail to capture dense surface information. Selecting a good point size and deciding the color of a rendered pixel in case of occlusion is not straightforward. Different methods have been proposed to address these issues. \\

\textbf{ Implicit representations can be viable alternatives to point clouds, voxels and meshes, but are computationally expensive when sampling points along a ray.} An implicit function describes a geometry at a 3D point by its occupancy probability, transparency, or distance to surfaces. A broad range of topologies can be represented at virtually infinite resolution and with low memory cost. Despite the advantages, aggregating data on a ray may be extensive to compute because sampling them requires evaluation of a neural network at a large number of points.

Table~\ref{table:dr_algorithms_comparison} shows the summary of the strength and limitations of different representations and rendering methods.

\subsection{Open Problems}
There are several problems that are not addressed by the current differentiable rendering methods.

Many applications such as 3D shape estimation and pose estimation, with the exception of 3D adversarial examples and style transfer, train neural networks by minimizing the difference between the rendered image and the target image. In such a pipeline, a rendering function and an image comparison function are usually developed separately. Therefore, a comparison function cannot leverage the rich 3D information that is fed into a rendering function. However, they can also be considered together. As mentioned in Section \ref{sec:algorithms:mesh}, the purpose of differentiable rendering is not to provide exact gradients, but to provide gradients that are useful for optimization. This could be achieved via a differentiable render-and-compare function that takes a 3D model and a target 2D image instead of a differentiable rendering function that takes a 3D model and outputs an image. One method in this direction is a differentiable projection and comparison of keypoints associated with the vertices of a 3D template shape~\cite{kanazawa2018learning}. Even though such a method is task-specific, it can be generalized by introducing the 3D information in the rendering and image comparison functions. The integration of such information is still an open research area.

Current differentiable rendering methods based on local illumination are very simple and cannot produce photorealistic images that exhibit shadows and reflections. Moreover, global illumination methods are too slow for training neural networks. On the other hand, in game engines, advances in real-time rendering methods have made it possible to render highly realistic images without using computationally-intensive path tracing. However, in the context of differentiable rendering, the area between these two is unexplored. Integrating fast, but complicated real-time rendering methods, such as shadow mapping and environment mapping, is an approach that has not been tried yet, but has the potential to improve the rendering quality.

Differentiable rendering of videos is also an interesting research direction to be explored. To achieve this, the integration of differentiable rendering with a physics simulator (to incorporate additional physical constraints) is needed. Theoretically, it should be possible to train an end-to-end pipeline that combines video data with differentiable physics simulators~\cite{de2018end,hu2019chainqueen}, but that has not been experimented yet.

Since rendering assumes a physical model to generate an image, the rendered image will appear unrealistic if there is a large discrepancy between the physical model and the real world. On the other hand, neural rendering can produce highly realistic images as it makes almost no assumptions about the physical model. However, such a lack of assumptions may sometimes produce images that violate the physical model. For example, in neural rendering, the shape of an object can vary depending on the viewpoint. The images of objects~\cite{nguyen2019hologan} and scenes~\cite{sitzmann2019scene} rendered by neural networks do not guarantee shape consistency across different viewpoints. While ways to incorporate inductive bias about the physical world into neural rendering are important, incorporating learning-based methods into differentiable rendering is also worth considering. Humans can intuitively and instantly understand how the scene will change when the camera moves. This ability is based on past experiences, rather than calculating the value of every pixel in the brain. Therefore, such kind of learning may facilitate gradient computation. Learning-based methods have already been used for image denoising in rendering~\cite{gharbi2019samplebased, kettunen2019} and efficient ray sampling~\cite{muller2019neural}. These approaches may also prove useful in differentiable rendering.

\begin{table*}[htp]
    \small
    \begin{center}
    \caption{Comparison between different algorithm types.}
    \label{table:dr_algorithms_comparison}
        \begin{tabular}{llll}
        \toprule
        Representation & Type & Strengths & Limitations \\
        \midrule
        \multirow{4}*{Mesh}& Analytical derivative & Usage of advanced forward rendering & Local minima in geometry optimization \\ 
        & Approximated gradient & Usage of advanced forward rendering & Handcrafting gradient calculation  \\ 
        & Approximated rendering & Auto-diff support & Not precise rendering result \\ 
        & Global illumination & Realistic rendering outcome & Computationally too expensive \\ \midrule
        \multirow{2}*{Voxel} & Occupancy / transparency & Simple, easy to optimize & Excessive memory consumption \\ 
        & Signed distance function & Efficient ray trace & Not suitable for transparent volume \\ 
        \midrule
        \multirow{2}*{Point Cloud} & Point cloud & Easy to render and differentiate &  Pseudo-sizing, lack of surface \\ 
        & RGBD image & Ordered point cloud by default & Point density reduces with the distance \\ 
        \midrule
        \multirow{2}*{Implicit} & Occupancy / transparency & Simple, easy to optimize & Inside and outside is ambiguous \\
        & Level set & Clear object boundary & Difficult to optimize \\ 
        \bottomrule
        \end{tabular}
    \end{center}
\end{table*}

\section{Evaluation}
\label{sec:evaluation}

Evaluation of differentiable rendering methods is not a trivial problem because rendering is a complex function. In this section, we review common practices for evaluating the algorithms and raise their problems.

A naive approach is direct gradient evaluation. For the global-illumination based methods~\cite{li2018differentiable,zhang2019differential,loubet2019reparameterizing,Zhang:2020:PSDR}, an efficient gradient computation method when the rendering integral includes visibility terms such as object boundaries is still an open research problem. Since the goal is computing analytically-correct gradients, gradients computed by finite differences are used as ground-truth. However, the lack of a common dataset for evaluation prevents algorithms for being evaluated quantitatively. On the other hand, some papers~\cite{loper2014opendr,kato2018neural,kato2019learning} focus on computing {\it approximated gradients} for local illumination models. In that case, the gradients should be meaningful for optimization rather then being analytically correct. For this reason, finite differences cannot be used for ground-truth. They only approximate the analytically correct derivative and, therefore, may not be practical for optimization. Loper and Black~\cite{loper2014opendr} compare gradients computed by their method with the ones computed by finite differences and claim that the proposed approach is better. Their reasoning is that determining a good epsilon in finite differences for optimization is challenging. Visualizing gradients (without showing ground-truth) and analyzing their convergence efficiency during the optimization of the objective function is another approach used for evaluation~\cite{kato2018neural,li2018differentiable}.

Evaluating the optimized scene parameters is another approach ~\cite{loper2014opendr,liu2019soft,chen2019dibrender} that is being employed. The lack of a common dataset for comparison leads each paper to use their own for evaluating the algorithm. Instead of optimizing the parameters of a 3D scene directly, many papers train neural networks for single-view 3D object reconstruction and report the reconstruction accuracy~\cite{kato2018neural,kato2019learning,yan2016perspective,tulsiani2017multi,liu2019soft,chen2019dibrender}. 

Computation time is also an important metric, especially for the ray tracing-based rendering methods. Therefore, several papers report computation time as part of the evaluation methodology~\cite{loper2014opendr,li2018differentiable,loubet2019reparameterizing}.

While evaluating the local and global illumination models, two main differences can be noticed. For the global ones, as the purpose of the algorithms is calculating analytically-correct gradients, the finite differences can be used as ground-truth. For the local ones, the purpose is to approximate useful gradients, therefore the optimization results should be used. However, in both cases, the lack of a common dataset prevents a fair comparison of the different algorithms. One possible solution is to create a set of toy problems which can be used to evaluate the derivative or optimization of geometry, material, light, and camera parameters. Hence, a novel evaluation methodology for differentiable rendering which is fair, accurate and meaningful is a necessity.

\section{Applications}
\label{sec:applications}

\begin{table}[t]
    \small
    \begin{center}
    \caption{Applications and representative literature of differentiable rendering.}
    \label{table:dr_applications}
    \begin{tabularx}{\linewidth}{lX}
        \toprule
        Application & Literature \\
        \midrule
        3D object reconstruction & \cite{kato2018neural,kato2019learning,yan2016perspective,tulsiani2017multi,tulsiani2018multi,liu2019soft,chen2019dibrender,petersen2019pix2vex,insafutdinov2018unsupervised,liu2019learning,niemeyer2019differentiable,kanazawa2018learning,li2020self,henderson2019learning,henzler2019escaping,yang2018learning,henderson2020leveraging,kundu20183d,liu2019dist} \\
        Body shape estimation & \cite{bogo2016keep,pavlakos2018learning} \\
        Hand shape estimation & \cite{baek2019pushing,zhang2019end,zimmermann2019freihand} \\
        Face reconstruction &  \cite{genova2018unsupervised} \\
        Object/camera pose estimation & \cite{rhodin2015versatile} \\
        Object part segmentation & \cite{deng2019cerberus} \\
        Material estimation & \cite{azinovic2019cvpr}\\
        Light/shading estimation & \cite{nieto2018robust,ramamoorthi2007first,khungurn2015matching,zhao2016downsampling} \\
        Adversarial examples & \cite{zeng2019adversarial,xiao2019cvpr,liu2018pixel,alcorn2019cvpr,wu2019sta} \\
        Auto-labeling & \cite{zakharov2019autolabeling} \\
        Teeth modeling & \cite{Velinov2018} \\
        \bottomrule
    \end{tabularx}
    \end{center}
\end{table}

Differentiable rendering has been used to tackle various 3D-related problems of computer vision and computer graphics. In this section we review applications of differentiable rendering and discuss their limitations. Table~\ref{table:dr_applications} shows representative applications of DR.

\subsection{Object Reconstruction}
\label{sec:applications:object-reconstruction}

\begin{figure*}[t]
    \begin{center}
        \includegraphics[width=0.9\linewidth]{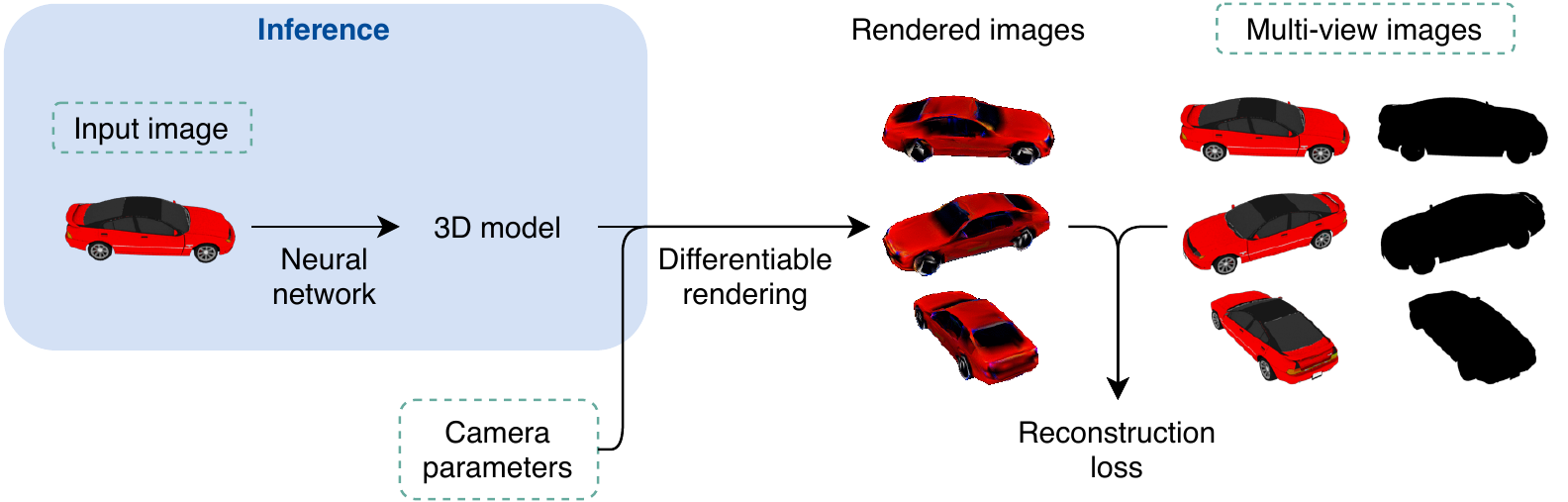}
        \caption{Standard training pipeline of learning single-view 3D object reconstruction from 2D images. Dashed rectangles represent training data.}
        \label{figure:object_reconstruction}
    \end{center}
\end{figure*}

\begin{figure}[t]
    \begin{center}
        \includegraphics[width=\linewidth,clip,trim=0 15.8cm 0 4.1cm]{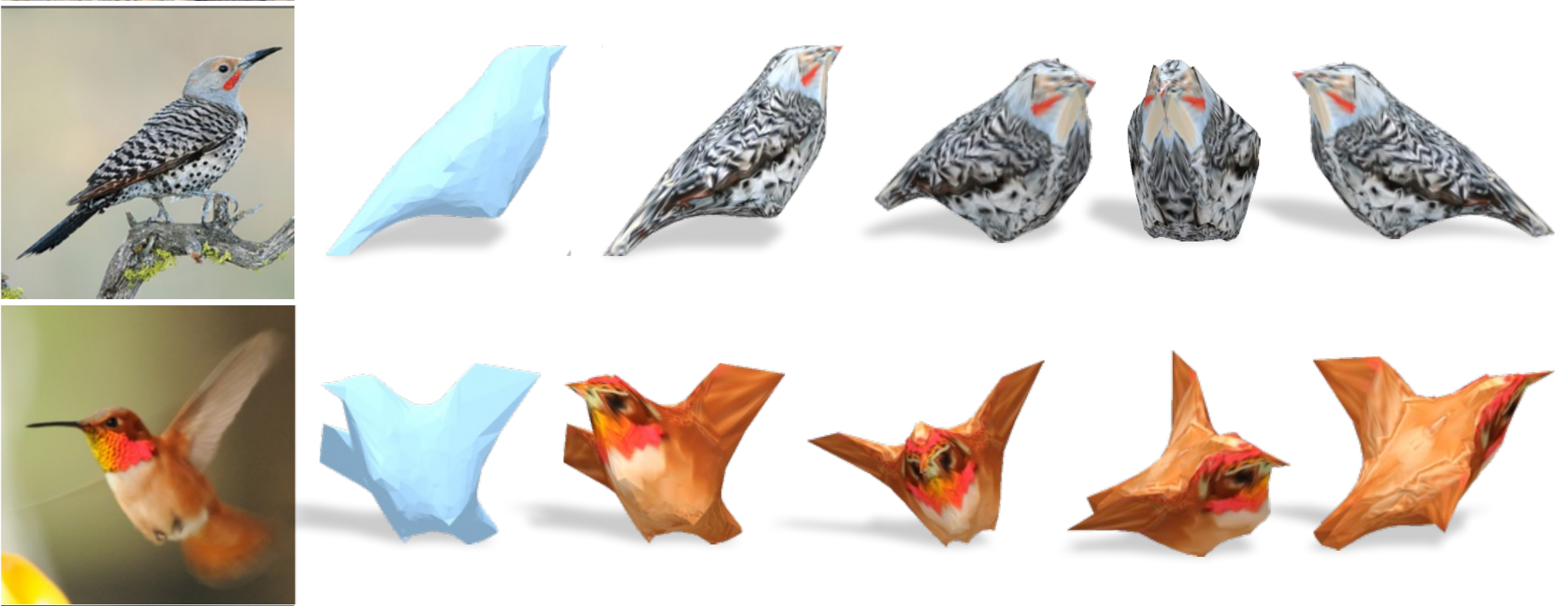}
    \end{center}
    \vspace{-5mm}
    \caption{Single-view 3D object reconstruction by Kanazawa \etal~\cite{kanazawa2018end}.} The leftmost column shows input images, and the rest are reconstructed objects.
    \label{figure:object_reconstruction_cmr}
\end{figure}

Single-view 3D object reconstruction is the task of estimating the 3D shape of an object from a single image. Unlike multi-view stereo and structure-from-motion, which use multiple images, this task is solved by machine learning rather than geometric estimation. Modern methods are able to learn high-quality 3D reconstruction from natural images, as shown in Figure~\ref{figure:object_reconstruction_cmr}.

A straightforward approach to single-view 3D object reconstruction is supervised learning using ground-truth 3D shapes~\cite{choy20163d,fan2017point,groueix2018papier,wang2020pixel2mesh,mescheder2019occupancy,chen2019learning}. Supervised learning, which requires the annotation of 3D shapes corresponding to images, is a costly process in terms of time and labor. Such a burden can be significantly reduced by replacing the 3D annotations with 2D ones. Self-supervision is commonly employed to achieve this. As differentiable rendering allows for observing the 3D scene parameters from the image space, the input image can be used to create supervision and help train a single view 3D object reconstruction pipeline, as shown in Figure~\ref{figure:object_reconstruction}.

An earlier work uses the non-differentiable OpenGL renderer for single-view 3D object reconstruction based on policy gradient algorithms~\cite{rezende2016unsupervised}. However, the set of the shapes to be reconstructed are limited to rough and simple ones. The advancement of voxel-based~\cite{yan2016perspective,tulsiani2017multi}, mesh-based~\cite{kato2018neural,liu2019soft,chen2019dibrender,petersen2019pix2vex}, point cloud-based~\cite{insafutdinov2018unsupervised} and neural implicit function-based~\cite{liu2019learning,niemeyer2019differentiable} differentiable rendering has significantly improved the reconstruction accuracy. Although the pipeline in Figure~\ref{figure:object_reconstruction} allows for learning a high quality 3D object reconstruction, it has two major issues. First, collecting multi-view, real-world images of an object is often impossible or too costly. Second, creating accurate annotations for some scene parameters, such as camera and light, is difficult for humans.

One possible solution for the first problem is to use a single image per object instead of multi-view images. However, due to its ill-posed nature, the reconstruction of the 3D shape becomes ambiguous when a single view is used during training. To account for this, Kanazawa \etal~\cite{kanazawa2018learning} propose to use the reprojection error of semantic keypoints that implicitly contain rough 3D information (e.g. beak of birds) as an additional training objective. Similarly, Liu \etal~\cite{li2020self} propose to use the reprojection error of semantic parts in a self-supervised manner. Although most methods do not model lighting effects, Henderson and Ferrari~\cite{henderson2019learning} explicitly provide light as an additional input to the renderer to use shading to help reduce the ambiguity of the reconstructed 3D shapes. Kato and Harada~\cite{kato2019learning} propose to use adversarial training in order to reconstruct shapes that looks realistic from any viewpoint.

For the second problem, several studies have attempted to eliminate or reduce the need for camera parameter annotations. Tulsiani \etal~\cite{tulsiani2018multi}, Insafutdinov and Dosovitskiy~\cite{insafutdinov2018unsupervised}, Henderson and Ferrari~\cite{henderson2019learning} propose to integrate a camera parameter estimation network into a shape estimation network and use the predicted parameters for rendering instead of using the ground-truth ones. The training objective of Henzler \etal~\cite{henzler2019escaping} does not focus on improving the reconstruction quality, but rather on being able to render realistic image from random viewpoints. On the other hand, Yang \etal~\cite{yang2018learning} propose a semi-supervised pipeline for learning the camera parameters.

Other topics in this field include learning a generative model of textured 3D objects~\cite{henderson2020leveraging}, reconstruction of cars in driving scenes by leveraging 3D CAD models~\cite{kundu20183d} and using differentiable rendering to fine-tune a shape estimation model trained with 3D supervision~\cite{liu2019dist}.

\subsubsection{Limitations and Open Problems}
Learning to reconstruct objects from natural image datasets of various categories is a possible research direction. Although not requiring 3D supervision is one of the advantages of learning from images, synthetic datasets are widely used in experiments. Such datasets are generated in a controlled environment in which objects are not occluded, image quality is very high and silhouette segmentation is almost noise-free. However, in real-world applications these assumptions often do not hold, which makes the synthetic dataset usage impractical for such scenarios. One of the commonly used synthetic dataset is ShapeNet~\cite{shapenet2015}, but other approaches ~\cite{kanazawa2018learning,li2020self,tulsiani2018multi,kato2019learning} use noisier natural image datasets such as Caltech-UCSD Birds-200-2011~\cite{wah2011caltech} and PASCAL~\cite{tulsiani2018multi,deng2009imagenet,everingham2010pascal,xiang2014beyond} for training. Due to inherent limitations of the available datasets, the trainable object categories are limited to bird, cars, aeroplanes and chairs, as key point annotations are required for camera parameter estimation. Novel approaches that rely on natural images will be the key to practical applications. Alternatively, learning from videos or from the interactions of scene elements is still a challenging problem. Humans also learn 3D reconstruction based on their observations, including the temporal changes and physical interactions. This type of supervision is less studied~\cite{novotny2017learning}, but would be an interesting topic to explore.

\subsection{Human Reconstruction}
\label{sec:applications:human-reconstruction}

\begin{figure}[t]
    \begin{center}
        \includegraphics[width=\linewidth,clip,trim=0 15.3cm 10cm 0]{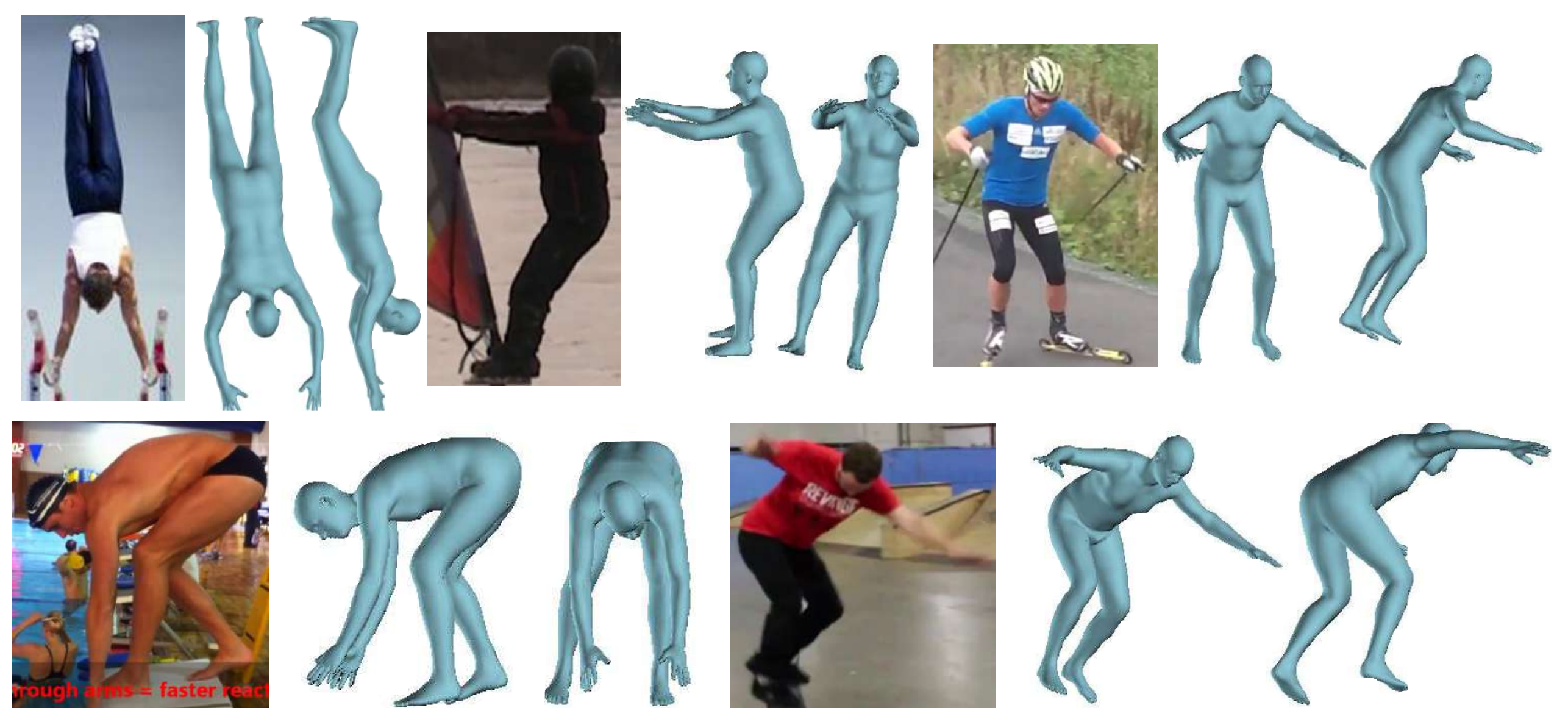}
        \vspace{-5mm}
        \caption{Human body estimation trained with silhouette and keypoint supervision by Pavlakos \etal~\cite{pavlakos2018learning} .}
        \label{figure:body_reconstruction} 
    \end{center}
    \vspace{-5mm}
\end{figure}

\subsubsection{Body Shape and Pose Reconstruction}
Human body shape and pose reconstruction is a problem that has long been studied by the vision community, since it opens up the possibility for a broad range of real-world applications. While remarkable success in this area has been obtained by processing information from alternative sensors~\cite{von2017sparse,weiss2011home}, image sequences~\cite{xu2018monoperfcap,alldieck2017optical} or multi-camera views~\cite{rhodin2016general,huang2017towards}, the progress has been limited for applications that only consider monocular images, due to the 2D to 3D mapping ambiguity. 

Recent advances in differentiable rendering allow for solving such ambiguities by learning prior knowledge that can be used to recover accurate 3D shapes. Most of the methods that we have surveyed employ a pipeline in which a statistical shape model is optimized for consistency between its projection to the image plane and key 2D image observations. Loper \etal~\cite{loper2015smpl} propose {\it SMPL}, a skinned vertex-based model that captures correlations in human shapes across the population, to address the full human body reconstruction problems. This model is a popular statistical body shape model and it can represent a large variety of body shapes, in natural human poses, with just a small set of parameters.

Bogo \etal~\cite{bogo2016keep} propose a method that predicts the pose and 3D mesh of the human body from a single image. The core observation is that body joints hold a rich set of information that can be exploited to recover the shape. Their method predicts joint locations in the 2D space using the method of Pischuilin \etal~\cite{pishchulin2016deepcut}. The body shape is modeled using SMPL and the pose is represented by a skeleton rig with 23 joints. Each joint encodes relative rotations between body parts. To recover the 3D shape and pose, the method minimizes an objective function that accounts for five terms: three pose priors that penalize unusual bending of knees and elbows, a shape prior that ensures consistency with the SMPL model and an error term that measures the distance between the projection of the SMPL joints and estimated ones in 2D. Kanazawa \etal~\cite{kanazawa2018end} further reconstruct human body shapes without using any 3D annotations.

The estimation of body shape and pose is also tackled by Pavlakos \etal~\cite{pavlakos2018learning} (Figure~\ref{figure:body_reconstruction}). They propose a pipeline in which a convolutional neural network is trained to predict both silhouettes and heatmaps corresponding to predefined keypoints. This information is used to learn pose and shape priors, which are fed into a body mesh generator (based on SMPL) that is trained to be consistent with the learned distributions. A differentiable renderer is further optimized for consistency between the mesh-to-image projection and the 2D annotations. Lassner \etal~\cite{lassner2017unite} similarly use silhouettes for reconstruction. Some recent works~\cite{natsume2019siclope,saito2019pifu} attempt to reconstruct human body shapes with clothes via a render-and-compare pipeline without using any statistical shape models and costly annotations.

Deng \etal~\cite{deng2019cerberus} tackle unsupervised 3D human body parts segmentation. The approach is based on a model that parametrizes a pre-defined number of parts by their relation to the camera and deformable triangular meshes. The neural network combines the parametrized parts into the same 3D space to obtain a 3D model of the whole object in a specific pose. Then, it optimizes for the unknown parameters using the render-and-compare approach. During inference, additional poses can be retrieved by sampling the parameter space in addition to a 3D model.

\subsubsection{Hand Shape and Pose Reconstruction}

Efforts have gone as well into accurate reconstruction of other parts of the body, such as hand shape and pose reconstruction. One of the early attempts to learn 3D hand poses from 2D images is the work of Oberweger \etal~\cite{oberweger2015training}. Their optimization pipeline relies on the synthetic depth rendering of the hand, which is achieved through a trained neural network. Differently, several works have tried to use differentiable rendering for this purpose. Baek \etal~\cite{baek2019pushing} address these challenges by proposing a parametric model that represents 3D deformable and articulated hand meshes. They train a feature extractor that incorporates texture and shape information of the hand, as well as a pose estimator, from color images. This information is further used to regress the vertices of a hand mesh and refine its pose, through an iterative optimization process. The authors also use the trained model to perform data augmentation by generating pairs of 3D meshes, corresponding to rendered 2D masks and RGB images. A similar approach is taken by Zhang \etal~\cite{zhang2019end}, in which heatmaps corresponding to 2D keypoints act as supervision signal for estimating the parameters a 3D hand model~\cite{romero2017embodied}. Qualitative experiments show that their method is able to accurately recover the hand shape and pose even in the presence of severe occlusions.

Zimmermann \etal~\cite{zimmermann2019freihand} report that hand shape and pose reconstruction methods often perform well on the datasets that they are trained on, but do not always generalize to other datasets or in real-world scenarios. They introduce a multi-view hand dataset accompanied by 3D pose and shape annotations, along with a methodology that allows for accurate annotation with moderate manual intervention. Their approach uses a sparse set of 2D keypoints and semi-automatically generated segmentation masks from multiple views by fitting a deformable hand model onto them. The fitting procedure yields a 3D hand pose and shape for each view, which are used to train a 3D hand pose estimation network. At inference time, the network predicts hand poses and corresponding confidence scores on unlabeled data. This allows a human annotator to save time by annotating the least confident predictions and verifying the remaining ones.

\begin{figure}[t]
    \begin{center}
        \includegraphics[width=0.13\linewidth]{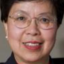}
        \includegraphics[width=0.13\linewidth]{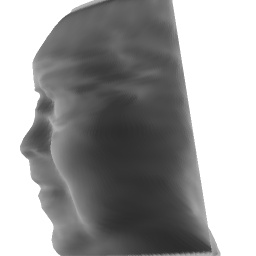}
        \includegraphics[width=0.13\linewidth]{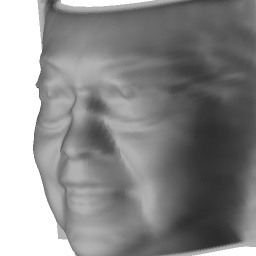}
        \includegraphics[width=0.13\linewidth]{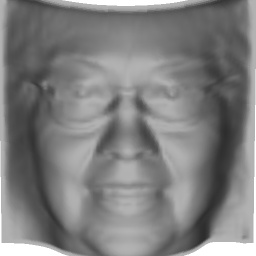}
        \includegraphics[width=0.13\linewidth]{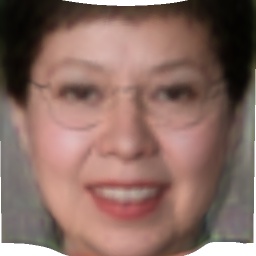}
        \includegraphics[width=0.13\linewidth]{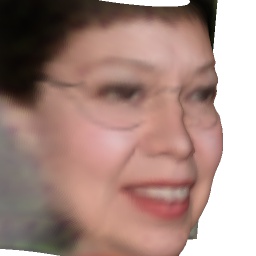}
        \includegraphics[width=0.13\linewidth]{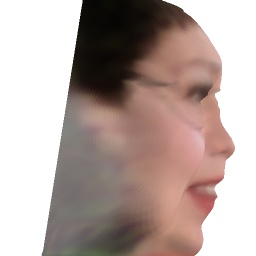} \\
        \includegraphics[width=0.13\linewidth]{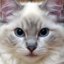}
        \includegraphics[width=0.13\linewidth]{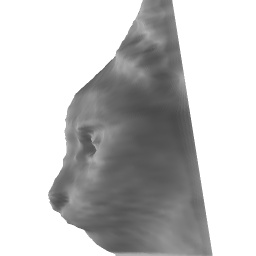}
        \includegraphics[width=0.13\linewidth]{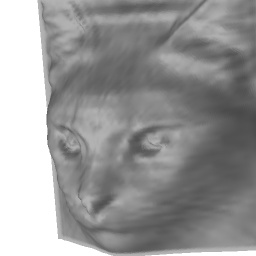}
        \includegraphics[width=0.13\linewidth]{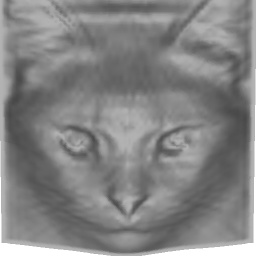}
        \includegraphics[width=0.13\linewidth]{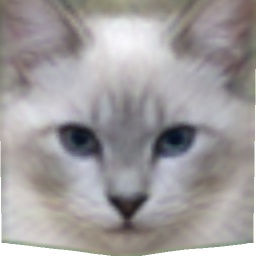}
        \includegraphics[width=0.13\linewidth]{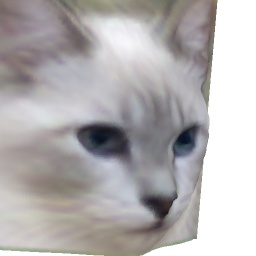}
        \includegraphics[width=0.13\linewidth]{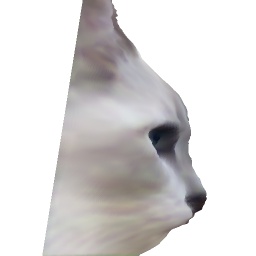} \\
        \vspace{-3mm}
        \caption{Unsupervised 3D face reconstruction by Wu \etal~\cite{wu2020unsupervised}.}
        \label{figure:face_reconstruction} 
    \end{center}
    \vspace{-5mm}
\end{figure}

\subsubsection{Face Reconstruction}
Other works have focused on face reconstruction, given the practical applications for the gaming and animation industries. Genova \etal~\cite{genova2018unsupervised} propose an unsupervised auto-encoder architecture which regresses the parameters of a textured 3D morphable model from image pixels. They leverage the features of a facial recognition network~\cite{schroff2015facenet} to regress the parameters of a statistical shape model of the face, using both real-world and synthetic face images. A differentiable renderer is used to preserve feature consistency across identities, to ensure that the parameter distributions within a batch remain consistent with the morphable model and that the network can correctly interpret its output. Wu \etal~\cite{wu2020unsupervised} propose a completely unsupervised method for learning face reconstruction from face images, which does not rely on existing 3D morphable face models (Figure \ref{figure:face_reconstruction}). Separate camera pose, depth, light and elbedo estimation networks learn to reconstruct the geomtry by leveraging shading information. A symmetry prediction network is used to reduce the 3D geometry ambiguity, assuming that most parts of shapes are symmetric.

\subsubsection{Limitations and Open Problems}

Most of the presented applications revolve around the reconstruction of a 3D shape from a single image. For human body shape reconstruction, the 2D to 3D mapping ambiguity can be alleviated using a statistical model~\cite{loper2015smpl,anguelov2005scape} to learn shape priors, whose parameters are optimized via differentiable rendering. However, existing models do not diversify age by including the infants, since they are trained on adult data. Hesse \etal~\cite{hesse2019learning} report that simply scaling the body model in order to fit to infant data does not produce the desired outcome, since body proportions are different. Finding ways to generalize across different age groups and body proportions is still an open research problem.

Existing body shape datasets are also either scarce or biased towards various body types and/or ethnicities. More variety in the data is required to improve the shape models, but collecting such data is challenging due to expensive 3D scanners, privacy issues and the difficulties for humans to stand still correctly during recording~\cite{hesse2019learning}.

Most of the body shape reconstruction methods that we have surveyed fit statistical shape models to 2D body joint locations, render the result and compare against the available observations. However, joint locations are not guaranteed to produce the optimal outcome and investigating alternative features should be considered.

Avoiding interpenetration is another challenge for human body reconstruction methods. Several works~\cite{gundogdu2019garnet,alldieck2017optical,pavlakos2019expressive} tackle this problem by introducing an error term in the objective function which penalizes unusual poses. However, this approach does not always avoid interpenetration, since the limbs are approximated with geometric primitives for efficiency purposes. Incorporating physics-based constraints in the rendering process to help solve this problem is still an open research area.

In the animation and gaming industry, producing realistic speech animations is a challenging task. As the lip-sync performance of an actor needs to be mapped to a virtual character, determining the accurate displacement of lips is crucial. Bhat \etal~\cite{bhat2013high} address this issue using helmet mounted cameras, which track face markers in order to find the blendshape weights of a face model. Since using markers is not always an option, other works~\cite{cao20133d,chen2013accurate,thies2016face2face} focus on using 2D features, depth or low-resolution images. Recently, muscle-based systems have shown potential to improve the accuracy and semantic interpretability of blendshape models. However, formulating and integrating such a model into a differentiable rendering-based pipeline is still an open research problem.

\subsection{3D Adversarial Examples}
\label{sec:applications:3d-adversarial}

Many machine learning models in computer vision are vulnerable to adversarial attacks, which are defined as inputs with visually imperceptible perturbations meant to intentionally trigger misclassification. Pixel perturbations are attractive due to their simplicity, but fail to consider the knowledge of image formation. This renders them unsuitable for real-world security applications, where an attacker does not have access to the pixel-level information. To account for these drawbacks, recent research work has been focusing on perturbing the object geometry and scene lighting parameter space.

Liu \etal~\cite{liu2018pixel} propose a method that perturbs scene lights, which are represented as spherical harmonics to filter unrealistic, high-frequency lighting while allowing usage of the analytical derivatives. On the other hand, Zeng \etal~\cite{zeng2019adversarial} perturb the intrinsic parameters of a scene/object. These changes in image intensities are constrained to be visually imperceptible for attacking a 2D object classifier. Xiao \etal~\cite{xiao2019cvpr} perturb object vertices with textures and rendered from various angles to study the vulnerable regions of the mesh. Alcorn \etal~\cite{alcorn2019cvpr} study attacks caused by out-of-distribution error by generating unrestricted 6D poses of 3D objects and analyzing how deep neural networks responded to such unusual configurations.

Adversarial attacks can also be applied to problems unrelated to image classification. Wu \etal~\cite{wu2019sta} analyze the vulnerabilities of Siamese networks~\cite{bertinetto2016fully} in the context of visual tracking. The authors claim that over-dependence of these trackers on similarity score, cosine window penalty and the asymmetrical region proposal network poses potential adversarial threats. They propose a differentiable rendering pipeline that generates perturbed texture maps for 3D objects, successfully lowering the tracking accuracy and even making the tracker drift off.

\subsubsection{Limitations and Open Problems}

Some applications are more vulnerable to adversarial attacks than others, due to limited training data being available or biased towards certain parts of the search space. For example, Alcorn \etal~\cite{alcorn2019cvpr} report that ImageNet classifiers only label 3.09\% of the 6D pose space of an object and misclassify many generated adversarial examples that are human-recognizable. As such, augmenting the training data using differentiable rendering with better photo-realistic reconstruction may help reduce the bias in the search space. Even though photo-realistic rendering has certain limitations (as explained in Section~\ref{sec:libraries:summary}), it is worth exploring the impact of data augmentation using differentiable rendering with simple illumination models.
\subsection{Other Applications}
\label{sec:applications:other-applications}

\begin{figure}[t]
    \begin{center}
        \includegraphics[width=\linewidth,clip,trim=7cm 1.9cm 7cm 7.4cm]{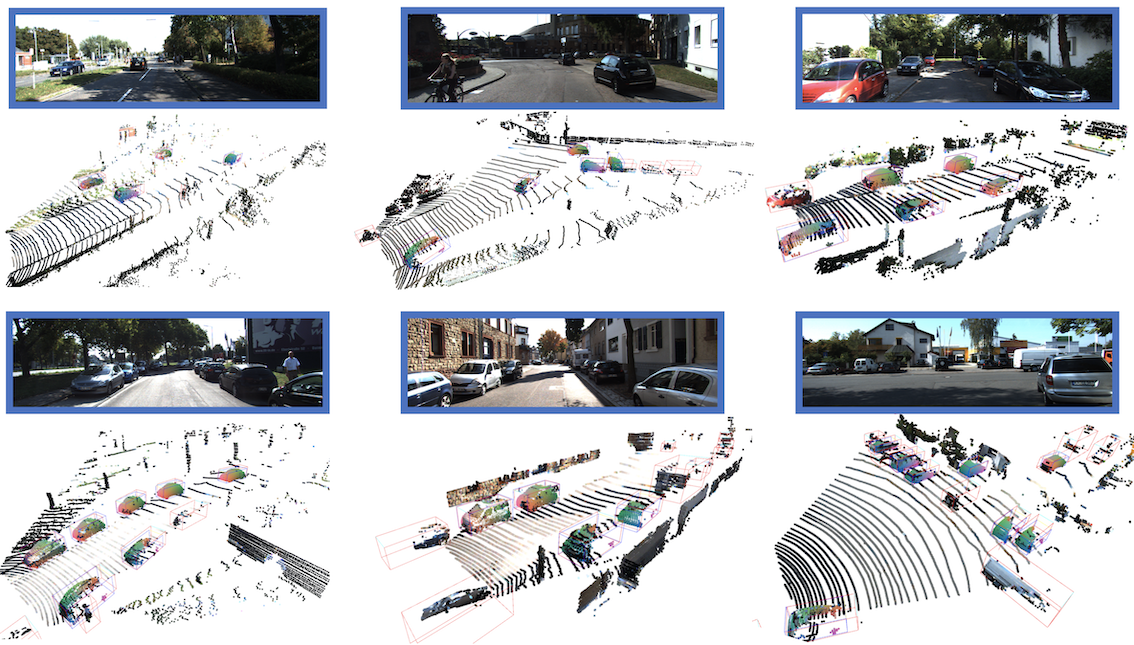} \\
        \includegraphics[width=\linewidth,height=0.367\linewidth]{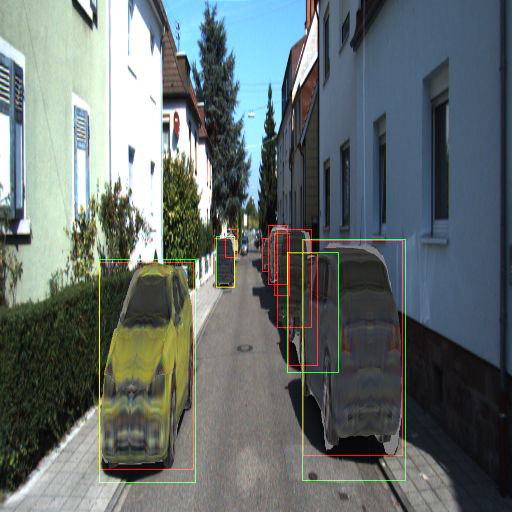}
        \vspace{-7mm}
        \caption{Automatic generation of 3D object detection labels via rendering by Zakharov \etal~\cite{zakharov2019autolabeling} (upper) and Beker \etal~\cite{beker2020self} (lower).}
        \label{figure:autolabeling} 
        \vspace{-1mm}
    \end{center}
\end{figure}

\begin{figure}[t]
    \centering
    \includegraphics[trim={86px 178px 18px 100px}, clip, width=0.19\linewidth]{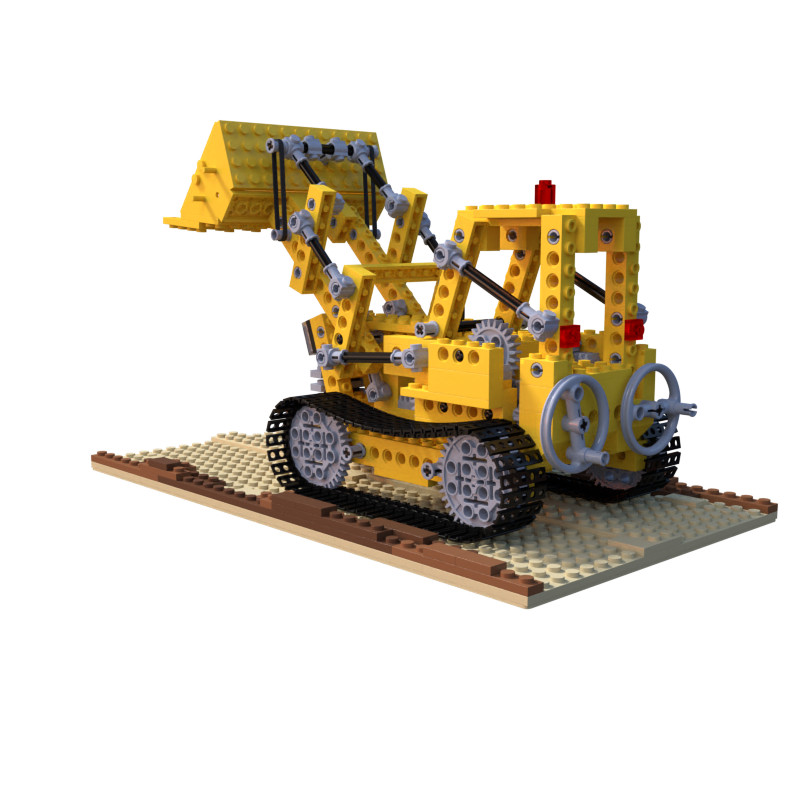}
    \includegraphics[trim={440px 420px 218px 238px}, clip, width=0.19\linewidth]{./sections/applications/figures/lego/gt.jpg}
    \includegraphics[trim={440px 420px 218px 238px}, clip, width=0.19\linewidth]{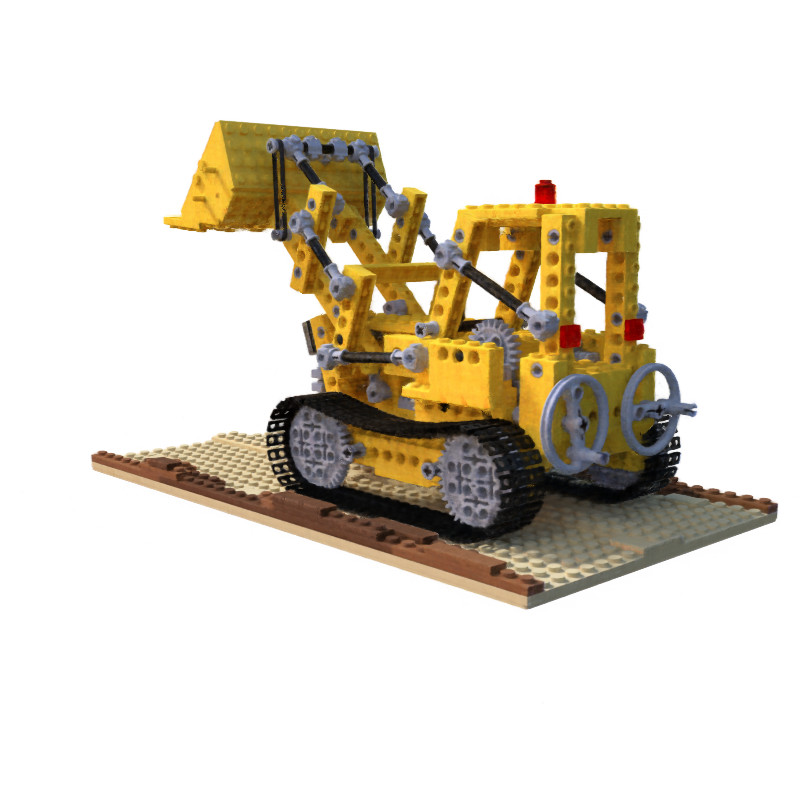}
    \includegraphics[trim={541px 289px 91px 343px}, clip, width=0.19\linewidth]{./sections/applications/figures/lego/gt.jpg}
    \includegraphics[trim={541px 289px 91px 343px}, clip, width=0.19\linewidth]{./sections/applications/figures/lego/nerf.jpg}
    \vspace{-3mm}
    \caption{Novel view syntheses by Mildenhall \etal~\cite{mildenhall2020nerf}. Ground-truth object (first image), cropped ground-truth (second and fourth), and cropped synthesized images (third and fifth).}
    \vspace{-2mm}
    \label{fig:nerf}
\end{figure}

\begin{figure}[t]
    \centering
    \includegraphics[width=0.8\linewidth]{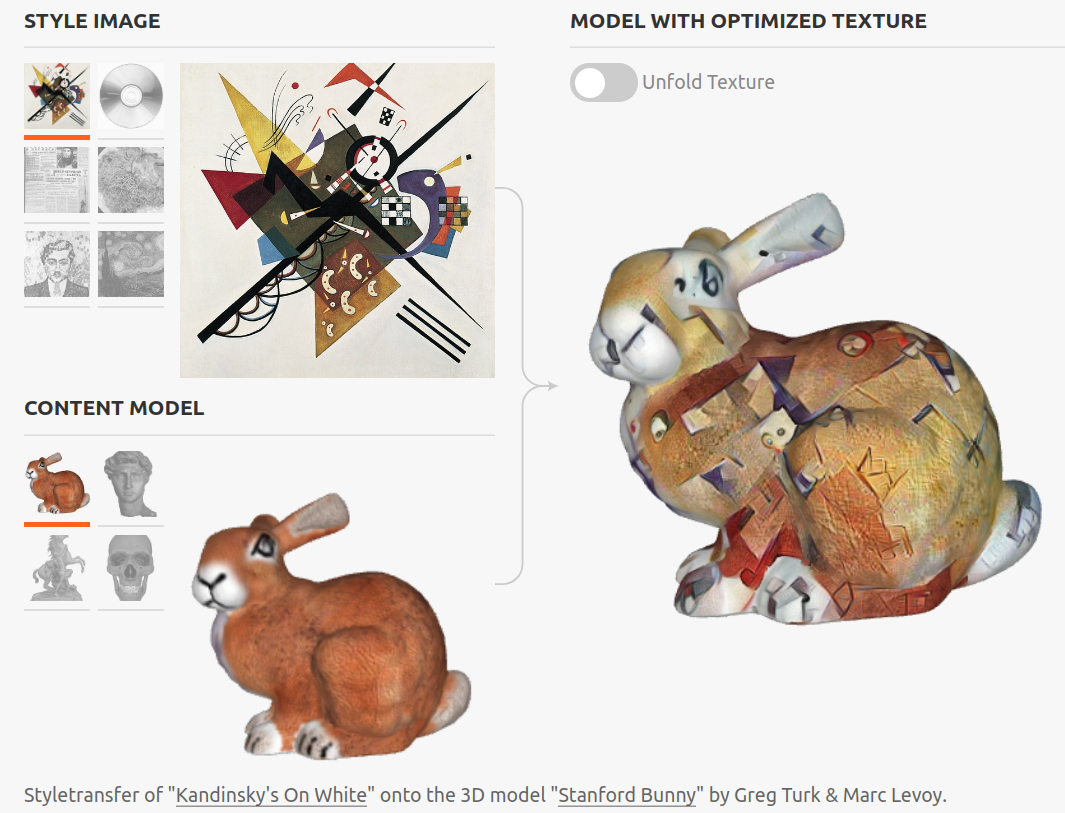}
    \vspace{-3mm}
    \caption{Style transfer of a 3D model by Mordvintsev \etal~\cite{mordvintsev2018differentiable}.}
    \label{fig:style_transfer}
\end{figure}

Although differentiable rendering has mainly been integrated into 3D object/human reconstruction and adversarial attacks pipelines, several works focus on different topics.

Zakharov \etal~\cite{zakharov2019autolabeling} propose an automatic annotation pipeline that recovers 9D pose (translation, rotation, scale) of cars from pre-trained off-the-shelf 2D detectors and sparse LiDAR data. They introduce a novel differentiable shape renderer based on SDF (signed distance fields) and NOCS (normalized object coordinate spaces)~\cite{wang2019cvpr}. The renderer optimizes for geometric and physical parameters, given the learned shape priors. Beker \etal~\cite{beker2020self} further improve accuracy (without using LiDAR information) by leveraging monocular depth estimation and textured 3D object reconstruction. Figure~\ref{figure:autolabeling} illustrates examples.

Several works focus on estimating the light sources from images. Nieto \etal~\cite{nieto2018robust} try to minimize the photo-metric error between the rendered image and the observed one, given the RGB and depth images of a scene. Unlike most previous works, which do not take cast shadows into account when estimating the illumination, their method is based on the Blinn-Phong reflection model~\cite{phong77}.

On the other hand, Azinovic \etal~\cite{azinovic2019cvpr} estimate the light source locations and object material properties in an indoor scene. They use 3D data and single or multiple RGB frames, along with corresponding camera poses, and employ a differentiable Monte Carlo renderer to optimize for the unknown parameters.

Differentiable rendering can also be integrated into a multi-view geometry pipeline to infer camera parameters and a dense surface model~\cite{smelyansky2002dramatic} or to synthesize novel views~\cite{mildenhall2020nerf} (Figure~\ref{fig:nerf}). In addition, differentiable path tracing is currently used to capture and model human teeth~\cite{Velinov2018}, to estimate shader parameters~\cite{khungurn2015matching,zhao2016downsampling},
to reconstruct objects which are not directly visible from the sensors~\cite{tsai2019beyond}, to jointly estimate the material and light parameters of a 3D scene~\cite{azinovic2019cvpr}, to estimate poses~\cite{rhodin2015versatile} and to analyze the lighting, shading and shadows in a scene~\cite{ramamoorthi2007first}.

Differentiable rendering is also useful for processing 3D models via rendered images~\cite{kato2018neural,liu2018paparazzi,mordvintsev2018differentiable} (Figure~\ref{fig:style_transfer}).

\subsubsection{Open Problems}

Several problems that have previously been studied could be improved using differentiable rendering. Examples include hand tracking~\cite{de2008model}, fabricating translucent materials~\cite{papas2013fabricating}, creating refractive caustics~\cite{schwartzburg2014high} or optimizing daylight as a part of an architectural design process~\cite{andersen2008intuitive}.

Image manipulation via 3D reconstruction is a novel application of differentiable rendering. Although initial works have been limited to driving scenes~\cite{yao20183d}, thanks to recent advancements in human and object reconstruction, other fields such as face image manipulation can be exploited by differentiable rendering.

\section{Libraries}
\label{sec:libraries}

In this section, we present existing differentiable rendering libraries. We also briefly cover the non-differentiable ones, as they have a relatively long history and can guide the future development of the differentiable ones. We discuss the limitations and open problems as well. 

We define a {\it differentiable rendering library} as the software that supports single or multiple differentiable rendering algorithms, implements multiple utility functions that can be used together with the rendering layer, can be integrated with existing neural network frameworks and is designed to be modular and extensible. We exclude an implementation of a single algorithm from the scope of the library definition.

\subsection{Differentiable Rendering Libraries}
\label{sec:libraries:differentiable-rendering}

As of second quarter of 2020, the differentiable rendering libraries available on the market are TensorFlow Graphics~\cite{TensorflowGraphicsIO2019}, Kaolin~\cite{kaolin2019arxiv}, PyTorch3D~\cite{ravi2020pytorch3d} and Mitsuba 2~\cite{nimier2019mitsuba}. We present a comparison of the library functionalities in Table ~\ref{table:dr_libraries_comparison}. Due to continuous development, there is a possibility for the list of functionalities to be extended by the time this paper is published.

\subsubsection{TensorFlow Graphics}
TensorFlow Graphics is being developed by Google and is the first library in this field. It combines differentiable rendering with the geometry layers under its scope to allow for easier integration with TensorFlow-based neural network architectures. However, it only supports analytical derivative of rendering described in Section~\ref{sec:algorithms:mesh} and does not provide gradients for visibility changes. The library supports the modeling of surface and material properties, the Phong reflectance model for Lambertian surfaces with point light and spherical harmonic lighting, graph convolution support and the Chamfer distance loss. It also allows meshes to be easily visualized in TensorBoard. In addition, the Levenberg-Marquardt algorithm~\cite{levenberg_marquardt} is provided for optimization. 

\subsubsection{Kaolin}
Kaolin is being developed by NVIDIA and is based on the PyTorch deep learning framework. Aside from differentiable rendering functionalities, the library includes implementations of various 3D neural architectures~\cite{chen2019dibrender, qi2016pointnet, qi2017pointnetplusplus, Zhang2019GRESNETGR, 3dgan, Wang_2018_ECCV, geometrics, mescheder2019occupancy} and support for triangle/quad mesh, point cloud, voxel grid and SDF representations of 3D objects. The neural architecture implementations are backed by a model zoo, which provides pretrained weights for each implementation. Kaolin preserves the original implementations of the existing differentiable rendering algorithms~\cite{kato2018neural,liu2019soft,chen2019dibrender} and provides support for the reading and preprocessing of various datasets~\cite{dai2017scannet,modelnet,shapenet2015}. For easier integration with different datasets and architectures, the library allows the conversion between various primitive types.

\subsubsection{PyTorch3D}
PyTorch3D is being developed by Facebook and is based on the PyTorch deep learning framework. The core rendering algorithm and all of its dependencies are optimized through CUDA implementations for both the backward and forward passes. The rendering pipeline is composed of a rasterizer and a shader module. The rasterizer applies the 3D camera transformations and rasterization to the given shape primitives, mesh and point cloud. The output is forwarded to the shader component which applies lighting, interpolates textures and calculates reflectance shadings including hard/soft Phong~\cite{phong77}, hard/soft Gouraud~\cite{gouraud1971}, hard flat and soft silhouette shading to create the final image. Besides the core rendering pipeline, PyTorch3D also supports several 3D loss functions including the Chamfer distance, mesh edge loss~\cite{Wang_2018_ECCV}, Laplacian regularization loss~\cite{Wang_2018_ECCV} and normal consistency loss~\cite{Wang_2018_ECCV}, along with graph convolutions. As a unique functionality, PyTorch3D can support heterogeneous batching of 3D data which allows usage of different sized meshes as input of the renderer. Such functionality is essential to represent 3D object efficiently with different level of details to find the balance between the vertex count and the visual quality depending on the application. 

\subsubsection{Mitsuba 2}
Mitsuba 2 proposes a novel approach for efficient high-performance rendering with auto-differentiation. It is designed as an efficient, internally portable for different compute units for multi-purpose usage. For optimization purposes, its Just-in-time (JIT) compiler supports symbolic execution of algorithms on GPU. Auto-differentiation is also supported through the Enoki library. Mitsuba 2 demonstrates its effectiveness and ease of use for tackling challenging problems: the Markov Chain Monte Carlo (MCMC) rendering technique is used to explore nearby light paths coherently to ensure faster convergence, a method for creating gradient-index optics which focuses incident illumination into caustics, and a method for reconstructing the parameters of a heterogeneous participating medium from multiple images.

Mitsuba 2, different from other libraries, focuses mainly on speed and efficiency through three guiding principles: no duplication, unobtrusiveness and modularity. It performs symbolic arithmetic, which delays the evaluation of a kernel until the variable is accessed. For the auto-differentiation-based algorithms, all the calculated variables and intermediate values are kept in memory. Especially for the inter-reflection scenes, such an approach consumes a significant amount of memory, which limits the rendered image size and the 3D object polygon count that can be processed at a time. To overcome this problem, Mitsuba 2 simplifies the computation graph periodically before each JIT compilation pass, using vertex elimination~\cite{osti_10118065,Yoshida1987Doa}.

Besides the efficient implementation and design, Mitsuba 2 proposes two novel MCMC schemes, {\it Coherent Pseudo-Marginal MLT} and {\it Multiple-Try Metropolis}. Coherent Pseudo-Marginal MLT algorithm is based on idea sampling a modified target function $\pi$ that is the result of convolving the path contribution function $f$ with a Gaussian kernel $G$. Such an approach produces coherent bundles of rays at each iteration and enhances the memory access speed assuming that the coherent ray's physical memory locations are close to each other. Multiple-Try Metropolis(MTM) is a modified version of the previous work, which integrates MTM into a vectorized MTL~\cite{Segovia2007CoherentML,Segovia:07:MIR}. The novel MTM algorithm generates a set of \textit{N} proposals to choose from at each iteration instead of the conventional random walk.

\subsubsection{Comparison}
All the libraries are optimized either through deep learning frameworks, which are well established for CPU/GPU agnostic coding with Python, or through custom C++/CUDA access. Mitsuba 2 is the only library that is based on a custom auto-differentiation module, Enoki. It also provides Python bindings on top of the C++ core, while other libraries benefit from the auto-differentiation capabilities of the PyTorch/TensorFlow libraries in Python. 

Tensorflow Graphics and Kaolin aim to integrate novel algorithms as 3rd party into their structure by preserving the original implementation. Such collection-oriented design choice reduces the required work of integrating the novel algorithms and allows easy comparison between different algorithms. However, as the rendering pipeline of each algorithm and their optimization levels are not centralized, the learning curve of a new algorithm, integration between multiple algorithms and the rendering speed are highly susceptible to vary significantly. On the other hand, PyTorch3D and Mitsuba 2 aim to provide a customizable pipeline where users can extend the capabilities of each sub-module of it. Such customization-oriented design choice allow novel algorithms to be integrated and developed using extensible components for lighting, shading, texturing and blending. While such design choice allows users to flexibly manipulate the renderer components and to experiment new research ideas efficiently, it requires an extra effort to integrate 3rd party algorithms if they are released as standalone implementations.

Besides the rendering capabilities of each library, Kaolin's ready to use state-of-art architecture implementations, along with pretrained model zoo, reduces the re-implementation and re-training cost. Also, having dataset loading and preprocessing functions is a strong advantage compared to other alternatives. On the other hand, the native support for the OBJ and PLY formats in PyTorch3D and Mitsuba 2 reduces the dependency on other libraries for I/O operations. The integration between TensorBoard and TensorFlow Graphics allows users to easily visualize rendered output. On the other hand, Kaolin provides several functions for visualizing meshes, voxels, and point clouds through the {\it pptk} backend. 
 
Besides Mitsuba 2, each library focuses on rasterization instead of ray tracing. This allows them to be easily usable with synthetic datasets and simple scenes without complex reflectance. Considering that the number of supported shading algorithms and surface materials, real world scene applications of existing libraries have so far been limited.

\begin{table*}[htp]
    \small
    \caption{Comparison between existing differentiable rendering libraries}
    \centering
    \begin{tabularx}{\linewidth}{lllll}
        \toprule
                                            & PyTorch 3D                  & Kaolin                      & TensorFlow Graphics               & Mitsuba 2 \\
        \midrule
        \multirow{2}*{Core implementation}  & \multirow{2}*{PyTorch/CUDA} & \multirow{2}*{PyTorch/CUDA} & \multirow{2}*{TensorFlow/C++/GPU} & C++/CUDA \\
                                            &                             &                             &                                   & Python bindings \\
        \midrule
        \multirow{4}*{Supported primitives} &                             & Triangle/quad mesh          &                                   & \\
                                            & Triangle mesh               & Point cloud                 & Triangle mesh                     & Triangle mesh \\
                                            & Point cloud                 & Voxel grid                  & Point cloud                       & Custom \\
                                            &                             & SDF                         &                                   & \\
        \midrule
        \multirow{3}{0.2\linewidth}{Differentiable rendering algorithms} & \multirow{3}{0.15\linewidth}{Soft Rasterizer (extendable)} & NMR                  & \multirow{3}*{Analytical derivatives} & \multirow{3}*{Loubet \etal} \\ 
                                                                         &                                                           & Soft Rasterizer      &                                       & \\
                                                                         &                                                           & DIB-Renderer         &                                       & \\
        \midrule
        Rendering Method                                                 & Rasterization                                             & Rasterization        & Rasterization                         & Ray tracing \\
        \midrule
        \multirow{4}*{3D operations}        & Graph convolutions         & Primitive conversions                                     & Graph convolutions   &  \\
                                            & 3D transformations         & 3D transformations                                        & 3D transformations   & 3D transformations \\
                                            & Point Cloud Operations         &                                         &   &  \\
                                            & (Umeyama, ICP, KNN)         &                                         &   &  \\
        \midrule
        \multirow{4}*{Shader support}       & Hard/soft Phong            & \multirow{4}*{Phong}                                      & \multirow{4}*{Phong} & \multirow{4}*{Wide range BSDF} \\
                                            & Hard/soft Gouraud          &                                                           &                      & \\
                                            & Hard flat                  &                                                           &                      & \\
                                            & Soft silhouette            &                                                           &                      & \\
        \midrule
        \multirow{6}*{Lighting support}     &                            &                                                           &                      & Area \\
                                            &                            &                                                           &                      & Point \\
                                            & Point                      & Ambient                                                   & Point                & Spot \\
                                            & Directional                & Directional                                               & Spherical harmonic   & Constant environment \\
                                            &                            &                                                           &                      & Environment map \\
                                            &                            &                                                           &                      & Directional \\
        \midrule
        \multirow{4}*{Loss functions}       & Chamfer Distance           & \multirow{4}*{\shortstack[l]{Chamfer distance \\ Directed distance \\ Mesh Laplacian}} & \multirow{4}*{Chamfer distance} & \multirow{4}*{Not supported} \\
                                            & Mesh edge                  &                                                           &                      & \\
                                            & Laplacian smoothing        &                                                           &                      & \\
                                            & Normal consistency         &                                                           &                      & \\
        \midrule
        \multirow{4}*{Camera support}       &                            &                         & \multirow{4}*{\shortstack[l]{Perspective \\ Ortographic \\ Quadratic distortion}} & Perspective \\
                                            & Perspective                & Perspective             &                                                                                   & (pinole, thin lens) \\
                                            & Ortographic                & Ortographic             &                                                                                   & Irradiance meter \\
                                            &                            &                         &                                                                                   & Radiance meter \\
        \midrule
        \multirow{2}*{Data I/O support}     & OBJ                        & \multirow{2}*{External} & \multirow{2}*{External} & OBJ \\
                                            & PLY                        &                         &                         & PLY \\
        \midrule
        \multirow{3}*{Data support}         & ShapeNet                     & ScanNet  & \multirow{3}*{Not supported} & \multirow{3}*{Not supported} \\
                                            & R2N2                         & ModelNet &                              & \\
                                            &                              & ShapeNet &                              & \\
        \midrule
        \multirow{8}*{Architectures}        & \multirow{8}*{Not supported} & DIB-Renderer       & \multirow{8}*{Not supported} & \multirow{8}*{Not supported} \\
                                            &                              & PointNet           &                              & \\
                                            &                              & PointNet++         &                              & \\
                                            &                              & GResNet            &                              & \\
                                            &                              & 3D-GAN             &                              & \\
                                            &                              & Pixel2Mesh         &                              & \\
                                            &                              & GEOMetrics         &                              & \\
                                            &                              & Occupancy Networks &                              & \\                                            
        \midrule
        \multirow{2}*{Other functionalities} & \multirow{2}*{Heterogeneous batches} & Model zoo     & \multirow{2}*{TensorBoard (mesh)} & Scene file support \\
                                             &                              & Visualization &                                   & Extensible by plugin \\
        \midrule
        Version                             & 0.2                        & 0.1.0              & 1.0.0                        & 2.0.0 \\ 
        \bottomrule
    \end{tabularx}
    \label{table:dr_libraries_comparison}
\end{table*}

\subsection{Non-Differentiable Rendering}
\label{sec:libraries:non-differentiable-rendering}

Several non-differentiable rasterizer-based and ray tracing-based rendering libraries have been developed over the last few decades. 

Rasterization-based libraries are preferred for applications where the rendering speed is more important than the visual quality. They commonly rely on the Direct3D~\cite{blythe2006direct3d}, OpenGL~\cite{OpenGLOv27:online} or Vulkan~\cite{VulkanOv49:online} APIs as backend instead of providing custom implementations. These APIs support common functionalities such as easy GPU access, ready-to-use rasterization pipeline and z-buffer algorithms. They implement their own shader language to allow developers to extend their capabilities. However, due to the large development cost incurred by the direct use of those APIs, it is also common to use the high-level wrappers Unity~\cite{UnityRea7:online} or Unreal Engine~\cite{UnrealEngine:online}. The game engines provide additional utility functions and allow designing interfaces. This allows easier portability of the rendering software across multiple hardware devices whose direct access API may be significantly different. It also reduces the knowledge requirement of a developer to hardware-specific content. Major strengths of high-level libraries are the easy integration of complex lights and shadows, animation functions, advanced I/O system for saving/loading various assets and scene information for reducing the research and development cost. The engines also support multiple programming languages for development, which helps reduce the learning curve required when writing production-level code. Lastly, they provide GUIs for easy interaction and visualization. This speeds up the development process and helps developers notice possible bugs early on, by visualizing intermediate results.

Photo-realistic rendering by ray tracing consumes more resources than rasterization. Hence, ray tracing-based libraries are preferred for applications where the visual quality is more important than the rendering speed. Arnold~\cite{ArnoldRe73:online,georgiev2018arnold}, V-Ray~\cite{ChaosGro23:online} and RenderMan~\cite{PixarsRe23:online, upstill1989renderman} are popular ray tracing-based libraries. They are integrated into digital content creation (DCC) tools such as Maya~\cite{MayaSoft95:online} and 3ds Max~\cite{3dsMax3D84:online} to provide end-user GUI access. For the materials and shaders, RenderMan provides its own shader language, RSL~\cite{upstill1989renderman}, as an extension to the editing capabilities through GUI. These libraries support voxel and parametric surfaces as shape primitives, in addition to meshes. Similar to rasterization, industrial ray tracing libraries are based on several low-level APIs such as Embree~\cite{wald2014embree}, OSPRay~\cite{wald2016ospray}, OptiX~\cite{parker2010optix} and Iray~\cite{keller2017iray}. Embree and OSPRay are optimized for Intel CPUs by using SIMD instructions. While Embree provides the ray tracing acceleration kernels, OSPRay contains extra functionalities such as volume rendering, global illumination, easier industry adoption, distributed computation and support for multiple shape primitives. On the other hand, the GPU-based libraries OptiX and Iray also follow the same structure. While OptiX provides the core kernel functionalities, Iray adds extra features like AI denoising, RTX hardware support, multi-GPU setup and easier third-party integration.

\subsection{Limitations and Open Problems}
\label{sec:libraries:summary}

Despite the emergent development of differentiable rendering libraries, there are still limitations and possible advancement directions to be addressed. We list some of them below:

\textbf{ Support for embedded environments is limited.} Optimizing an algorithm is essential to enable it to run on embedded environments where hardware resources and processing power are limited. For NVIDIA GPUs, TensorRT\footnote{https://developer.nvidia.com/tensorrt} has been developed for optimizing a trained network's inference speed. However, a similar functionality for optimizing differentiable rendering algorithms for embedded environments is still limited. 

\textbf{ Current libraries do not provide a standard interface for extensions.} Current differentiable rendering libraries either require new algorithms to be implemented from scratch or provide basic extension layers through a plugin-based architecture. Providing extension capabilities which are common across the frameworks, similar to the non-differentiable rendering community, would be an added value.

\textbf{ Existing implementations have limited functionalities.} Differentiable rendering libraries are designed mainly for computer vision researchers and engineers. Functionalities that currently exist in the non-differentiable ones, such as integration with DCC tools, support for primitives other than triangles, various output formats such as high-dynamic range images, meta-data/animation rendering, are currently lacking. We believe that adding more functionalities required for the game and movie industries would enable the development of new types of applications. Also, existing libraries (except for Kaolin) do not provide ready-to-use support for the datasets which are commonly used in differentiable rendering research. This, in turn, leads to repetitive implementation of dataset management, while increasing the development costs. 

\textbf{ Support for light and material models is limited.} As light and surface material quality are essential for photo-realistic RGB rendering, the development of advanced light support for both rasterization and ray tracing-based libraries is limited and not unified. Such limitations of the light models hinder the realistic appearance of the rendered RGB image and, therefore, the quality of the supervision signal.

\textbf{ Limited debugging and benchmarking support.} Debugging tools and benchmarking support are crucial for efficient development and research. Existing libraries benefit from language-specific and/or third-party debugging tools for benchmarking, if any. A benchmarking and debugging tool specifically designed for differentiable rendering, which aims to detect both rendering and gradient calculation inaccuracies, is yet to be developed. 

\textbf{ Model sharing across different libraries is currently not possible.} Support for model sharing across multiple deep learning libraries is another limitation. Recently, the ONNX format\footnote{\url{https://onnx.ai}} has been developed to allow for easy sharing of trained models. However, due to the implementation differences across the different libraries, it is still not possible to export a neural network model which contains differentiable rendering layers and use it in a different library for inference through the ONNX interface.

\section{Conclusion}
\label{sec:conclusion}

This paper presented an overview of the current state of differentiable rendering research - popular algorithms and data representation techniques, as well as evaluation metrics and common practices. It also introduced applications that make use of DR-based techniques, covering a broad range of fields such as 3D object/body shape and pose reconstruction, adversarial attacks, auto labeling or light source estimation. Several libraries that are commonly used to develop new differentiable rendering-based methods were presented and compared. Finally, open problems for algorithms, applications and libraries were discussed.

Although differentiable rendering is a novel field, it is rapidly maturing, aided by the continuous development of new tools meant to simplify its usage. This will enable more researchers to develop novel neural network models that understand 3D information from 2D image data. Consequently, the performance of various applications can be improved, while the need for 3D data collection and annotation can be reduced. Once mature enough to be deployed to embedded devices, we will likely start seeing differentiable rendering-based methods solving problems with real-time constraints (such as autonomous driving). The times that we live are exciting and the advancements in deep neural networks help us tackle and solve everyday problems. The best is yet to come.
\label{sec:acknowledgements}
\ifCLASSOPTIONcompsoc
  \section*{Acknowledgments}
\else
  \section*{Acknowledgment}
\fi

This work was supported by Toyota Research Institute Advanced Development, Inc. We would like to thank the PyTorch3D developers for their insightful comments and suggestions. 

\ifCLASSOPTIONcaptionsoff
  \newpage
\fi

\bibliographystyle{IEEEtran}
\bibliography{IEEEabrv,references.bib}

\label{sec:biography}

\begin{IEEEbiography}[{\includegraphics[width=1in,height=1.25in,clip,keepaspectratio]{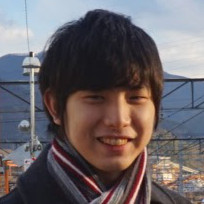}}]{Hiroharu Kato}
received his BS degree (2012) in Engineering and his MS degree (2014) in Information Science and Technology from the University of Tokyo, Japan. After spending two and a half years at Sony Corporation as a research engineer, he is currently a researcher at Preferred Networks, Inc. He is also a Ph.D. student at the University of Tokyo since 2016.
\end{IEEEbiography}

\vspace{-16mm}
\begin{IEEEbiography}[{\includegraphics[width=1in,height=1.25in,clip,keepaspectratio]{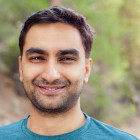}}]{Deniz Beker}
obtained his BSc. degree from Isık University in 2011 and his MSc. degree from Yeditepe University in 2014 both in Electrical and Electronics Engineering. He currently works at Preferred Networks as an engineer and engineering manager. His research interests are 3D computer vision and autonomous driving problems.
\end{IEEEbiography}

\vspace{-16mm}
\begin{IEEEbiography}[{\includegraphics[width=1in,height=1.25in,clip,keepaspectratio]{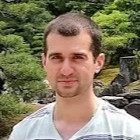}}]{Mihai Morariu}
obtained a BSc. (Mathematics) degree from the University of Bucharest (Romania) in 2011 and a MSc. (Artificial Intelligence) from the Univesity of Amsterdam (The Netherlands) in 2013. After having worked for several years in The Netherlands as a computer vision engineer (in robotics), he relocated to Tokyo in 2018. He is now working as an engineer for Preferred Networks, Inc.
\end{IEEEbiography}

\vspace{-16mm}
\begin{IEEEbiography}[{\includegraphics[width=1in,height=1.25in,clip,keepaspectratio]{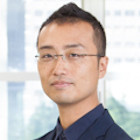}}]{Takahiro Ando}
obtained his BSc. degree from Tsukuba University in 2002. He currently works at Preferred Networks as engineer. His research interests are 3D computer vision and computer graphics.
\end{IEEEbiography}

\vspace{-16mm}
\begin{IEEEbiography}
[{\includegraphics[width=1in,height=1.25in,clip,keepaspectratio]{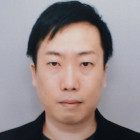}}]{Toru Matsuoka}
obtained his MSc. degree in Computer Science from Takushoku University in 2004. He started his career as a CAD/CAM software developer in 2005. After that, he joined Preferred Networks as an engineer in 2017 after having worked for a video production company and a game company. 
\end{IEEEbiography}

\vspace{-16mm}
\begin{IEEEbiography}[{\includegraphics[width=1in,height=1.25in,clip,keepaspectratio]{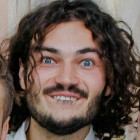}}]{Wadim Kehl}
is an ML engineer at the Toyota Research Institute - Advanced Development (TRI-AD) in Tokyo, Japan. His research focuses on the interface between 3D object perception and differentiable visual representations, and how to apply these insights to practical problems. He received his BSc. from the University of Bonn, and his MSc. as well as PhD from the Technical University of Munich. He has authored over 20 patents and publications, presented at CVPR/ICCV/ECCV, and is a reviewer for most top-tier conferences and journals in Computer Vision and ML.  
\end{IEEEbiography}

\vspace{-12mm}
\begin{IEEEbiography}[{\includegraphics[width=1in,height=1.25in,clip,keepaspectratio]{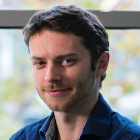}}]{Adrien Gaidon}
is the Head of Machine Learning Research at the Toyota Research Institute (TRI) in Los Altos, CA, USA. Adrien's research focuses on scaling up ML for robot autonomy, spanning Scene and Behavior Understanding, Simulation for Deep Learning, 3D Computer Vision, and Self-Supervised Learning. He received his PhD from Microsoft Research - Inria Paris in 2012, has over 40 publications and patents in ML/CV/AI, top entries in international Computer Vision competitions, multiple best reviewer awards, international press coverage for his work on Deep Learning with simulation, was a guest editor for the International Journal of Computer Vision, and co-organized multiple workshops on Autonomous Driving at CVPR/ECCV/ICML.
\end{IEEEbiography}

\end{document}